\documentclass[10pt,twocolumn,letterpaper]{article}

\usepackage{cvpr}
\usepackage{times}
\usepackage{epsfig}
\usepackage{graphicx}
\usepackage{amsmath}
\usepackage{amssymb}
\usepackage{float}

\usepackage[pagebackref=true,breaklinks=true,letterpaper=true,colorlinks,bookmarks=false]{hyperref}
\hypersetup{linkcolor=,urlcolor=}

 \cvprfinalcopy %

\def\cvprPaperID{7361} %

\usepackage[export]{adjustbox}
\usepackage{listings}
\usepackage{multirow}
\usepackage{makecell}
\usepackage{placeins}
\usepackage{colortbl}
\usepackage{rotating}
\usepackage{array}
\usepackage{color}
\usepackage{xcolor}
\usepackage{verbatim}
\usepackage{units}
\usepackage{textcomp}
\usepackage{bbding}
\usepackage{mwe}
\usepackage{subcaption}
\usepackage{comment}
\PassOptionsToPackage{normalem}{ulem}
\usepackage{ulem}
\usepackage{anyfontsize}

\definecolor{lightgray}{gray}{0.4}
\definecolor{verylightgray}{gray}{0.7}
\definecolor{veryverylightgray}{gray}{0.9}

\definecolor{darkgreen}{rgb}{0, 0.5, 0}

\makeatletter

\renewcommand{\paragraph}{%
	\@startsection{paragraph}{4}%
	{\z@}{0.5ex plus 0.8ex minus .5ex}{-0.5em}{\normalsize\bf}}

\let\originalparagraph\paragraph
\renewcommand{\paragraph}[2][.]{\originalparagraph{#2#1}}
\makeatother

\newcommand{\beginsupplement}{%
	\setcounter{table}{0}
	\renewcommand{\thetable}{S\arabic{table}}%
	\setcounter{figure}{0}
	\renewcommand{\thefigure}{S\arabic{figure}}%
	\setcounter{section}{0}
	\renewcommand{\thesection}{S\arabic{section}}  
	\setcounter{equation}{0}
	\renewcommand{\theequation}{S\arabic{equation}}
}

\makeatletter

\begin{document}

\title{\vspace{-3em}A U-Net Based Discriminator for Generative Adversarial Networks}
\author{%
	Edgar Sch{\"o}nfeld\\
	Bosch Center for Artificial Intelligence\\
	\texttt{edgar.schoenfeld@bosch.com} \\
	\and
	Bernt Schiele\\
	Max Planck Institute for Informatics\\
	\texttt{schiele@mpi-inf.mpg.com} 
	\and
	Anna Khoreva \\
	Bosch Center for Artificial Intelligence\\
	\texttt{anna.khoreva@bosch.com} \\
}

\maketitle

\begin{abstract}
\vspace{-1.0em}
Among the major remaining challenges for generative adversarial networks (GANs) is the capacity to synthesize globally and locally coherent images with object shapes and textures indistinguishable from real images. To target this issue we propose an alternative U-Net based discriminator architecture, borrowing the insights from the segmentation literature. The proposed U-Net based architecture allows to provide detailed per-pixel feedback to the generator while maintaining the global coherence of synthesized images, by providing the global image feedback as well.
Empowered by the per-pixel response of the discriminator,
we further propose a per-pixel consistency regularization technique based on the CutMix data augmentation, encouraging the U-Net discriminator to focus more on
semantic and structural changes between real and fake images.
This improves the U-Net discriminator training, further enhancing the quality of generated samples.
The novel discriminator improves over the state of the art in terms of the standard distribution and image quality metrics, enabling the generator to synthesize images with varying structure, appearance and levels of detail, maintaining global and local realism. 
Compared to the BigGAN baseline, we achieve an average improvement of $2.7$ FID points across FFHQ, CelebA, and the newly introduced COCO-Animals dataset. The code is available at \url{https://github.com/boschresearch/unetgan}.

\end{abstract}

\vspace{-0.5em}
\section{Introduction}\label{intro} 
\vspace{-0.5em}
The quality of synthetic images produced by generative adversarial networks (GANs) has seen tremendous improvement recently \cite{Brock2019, Karras2018ASG}.
The progress is attributed to large-scale training \cite{Menick2018GeneratingHF,Brock2019}, architectural modifications \cite{Zhang_SAGAN19, karras2018progressive, Karras2018ASG, Lin2019COCOGANGB}, and improved training stability via the use of different regularization techniques \cite{miyato2018spectral, Zhang2019ConsistencyRF}.
However, despite the recent advances, learning to synthesize images with global semantic coherence, long-range structure and the exactness of detail remains challenging. 

\begin{figure}[t]
	\begin{centering}
		\setlength{\tabcolsep}{0.1em}
		\renewcommand{\arraystretch}{0}
		\par\end{centering}
	\begin{centering}
\begin{tabular}{@{\hskip -0.05in}c@{}}
	  \vspace{-0.03in}
	  Progression during training\\
    \includegraphics[width=0.935\columnwidth]{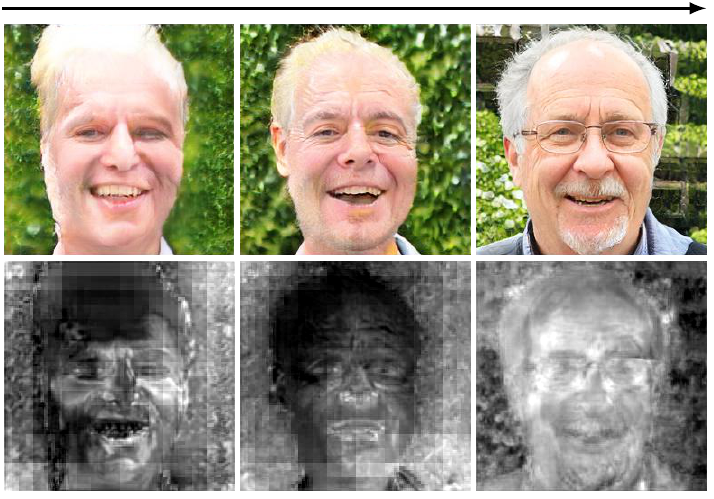}
	\includegraphics[width=0.065\columnwidth]{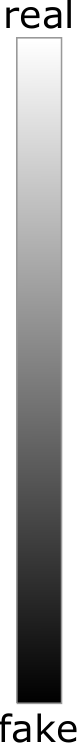}  
    	\end{tabular}
	
		\par\end{centering}
	\caption{\label{fig:teaser}Images produced throughout the training by our U-Net GAN model (top row) and their corresponding per-pixel feedback of the \textit{U-Net discriminator} (bottom row). The synthetic image samples are obtained from a fixed noise vector at different training iterations. Brighter colors correspond to the discriminator confidence of pixel being real (and darker of being fake). Note that the U-Net discriminator provides very detailed and spatially coherent response to the generator, enabling it to further improve the image quality, e.g. the unnaturally large man's forehead is recognized as fake by the discriminator and is corrected by the generator throughout the training.}
\end{figure}

One source of the problem lies potentially in the discriminator network. The discriminator aims to model the data distribution, acting as a loss function to provide the generator a learning signal to synthesize realistic image samples. 
The stronger the discriminator is, the better the generator has to become.
In the current state-of-the-art GAN models, the discriminator being a classification network learns only a representation that allows to efficiently penalize the generator based on the most discriminative difference between real and synthetic images. Thus, it often focuses either on the global structure or local details. The problem amplifies as the discriminator has to learn in a non-stationary environment: the distribution of synthetic samples shifts as the generator constantly changes through training, and is prone to forgetting previous tasks \cite{ChenSS2019} (in the context of the discriminator training, learning semantics, structures, and textures can be considered different tasks). 
This discriminator is not incentivized to maintain a more powerful data representation, learning both global and local image differences. This often results in the generated images with discontinued and mottled local structures \cite{Lin2019COCOGANGB} or images with incoherent geometric and structural patterns (e.g. asymmetric faces or animals with missing legs) \cite{Zhang_SAGAN19}. %

To mitigate this problem, we propose an alternative discriminator architecture, which outputs simultaneously both global (over the whole image) and local (per-pixel) decision of the image belonging to either the real or fake class, see Figure~\ref{fig:teaser}. Motivated by the ideas from the segmentation literature, we re-design the discriminator to take a role of both a classifier and segmenter. We change the architecture of the discriminator network to a U-Net  \cite{Ronneberger2015UNetCN}, where the encoder module performs per-image classification, as in the standard GAN setting, and the decoder module outputs per-pixel class decision, providing spatially coherent feedback to the generator, see Figure~\ref{fig:method_overview}.
This architectural change leads to a stronger discriminator, which is encouraged to maintain a more powerful data representation, making the generator task of fooling the discriminator more challenging and thus improving the quality of generated samples (as also reflected in the generator and discriminator loss behavior in Figure \ref{fig:g_d_loss}). Note that we do not modify the generator in any way, and our work is orthogonal to the ongoing research on architectural changes of the generator \cite{Karras2018ASG, Lin2019COCOGANGB}, divergence measures \cite{Li2017MMDGT, Arjovsky2017WGAN, Nowozin2016fGANTG}, and regularizations \cite{Roth_NeurIPS2017,gulrajani_NeurIPS2017,miyato2018spectral}.

The proposed U-Net based discriminator allows to employ the recently introduced CutMix \cite{Yun2019CutMixRS} augmentation, which is shown to be effective for classification networks, for consistency regularization in the two-dimensional output space of the decoder.
Inspired by \cite{Yun2019CutMixRS}, we cut and mix the patches from real and synthetic images together, where the ground truth label maps are spatially combined with respect to the real and fake patch class for the segmenter (U-Net decoder) and the class labels are set to fake for the classifier (U-Net encoder), as globally the CutMix image should be recognized as fake, see Figure~\ref{fig:cutmix}. Empowered by per-pixel feedback of the U-Net discriminator, we further employ these CutMix images for consistency regularization, penalizing per-pixel inconsistent predictions of the discriminator under the CutMix transformations. This fosters the discriminator to focus more on semantic and structural changes between real and fake images and to attend less to domain-preserving perturbations. Moreover, it also helps to improve the localization ability of the decoder. Employing the proposed consistency regularization leads to a stronger generator, which pays more attention to local and global image realism. We call our model U-Net GAN.

We evaluate the proposed U-Net GAN model across several datasets using the state-of-the-art BigGAN model~\cite{Brock2019} as a baseline and observe an improved quality of the generated samples in terms of the FID and IS metrics.
For unconditional image synthesis on FFHQ~\cite{Karras2018ASG} at resolution $256\times256$, our U-Net GAN model improves $4$ FID points over the BigGAN model, synthesizing high quality human faces (see Figure \ref{fig:qual_results}). On CelebA~\cite{Liu_Celeba} at resolution $128\times128$ we achieve $1.6$ point FID gain, yielding to the best of our knowledge the lowest known FID score of $2.95$.
For class-conditional image synthesis on the introduced COCO-Animals dataset~\cite{Lin2014MicrosoftCC,OpenImages} at resolution $128\times128$ we observe an improvement in FID from $16.37$ to $13.73$, synthesizing diverse images of different animal classes (see Figure~\ref{fig:coco_pics}).

\section{Related work}\label{related_work}

\paragraph{Generative adversarial networks}

GAN~\cite{goodfellow2014generative} and its conditional variant~\cite{Mirza2014ConditionalGA} have recently demonstrated impressive results on different computer vision tasks, including image synthesis \cite{Radford2016UnsupervisedRL,Zhang_SAGAN19,karras2018progressive, Brock2019, Karras2018ASG, Lin2019COCOGANGB,Dey2018RankGANAM}. %
Plenty of efforts have been made to improve the training and performance of GANs, from reformulation of the objective function \cite{MaoLXLW16,Arjovsky2017WGAN,Lim2017GeometricG,Nowozin2016fGANTG}, integration of different regularization techniques \cite{Zhang2019ConsistencyRF,miyato2018spectral,Roth_NeurIPS2017,Zhang2018PAGANIG} and architectural changes \cite{Radford2016UnsupervisedRL,karras2018progressive,Durugkar2016GenerativeMN,Lin2019COCOGANGB}.
To enhance the quality of generated samples, \cite{Radford2016UnsupervisedRL} introduced the DCGAN architecture that employs strided and transposed convolutions. In SAGAN \cite{Zhang_SAGAN19} the self-attention block was added to improve the network ability to model global structure. PG-GAN \cite{karras2018progressive} proposed to grow both the generator and discriminator networks to increase the resolution of generated images. Other lines of work focused mainly on improving the discriminator by exploiting multiple \cite{Mordido2018DropoutGANLF, Durugkar2016GenerativeMN, Doan2018OnlineAC} and multi-resolution \cite{Wang2017HighResolutionIS,Sharma2018ImprovedTW} discriminators, using spatial feedback of the discriminator \cite{Huh2019FeedbackAL}, an auto-encoder architecture with the reconstruction-based feedback to the generator~\cite{zhao2016energy} or self-supervision to avoid catastrophic forgetting \cite{ChenSS2019}. 
Most recently, the attention has been switched back to the generator network. StyleGAN \cite{Karras2018ASG} proposed to alter the generator architecture by injecting latent codes to each convolution layer, thus allowing more control over the image synthesis process. COCO-GAN~\cite{Lin2019COCOGANGB} integrated the conditional coordination mechanism into the generator, making image synthesis highly parallelizable. 
In this paper, we propose to alter the discriminator network to a U-Net based architecture, empowering the discriminator to capture better both global and local structures, enabled by per-pixel discriminator feedback. 
Local discriminator feedback is also commonly applied through PatchGAN discriminators~\cite{isola2017image}. Our U-Net GAN extends this idea to dense prediction over the whole image plane, with visual information being integrated over up- and down-sampling pathways and through the encoder-decoder skip connections, without trading off local over global realism.

\paragraph{Mix\&Cut regularizations}
Recently, a few simple yet effective regularization techniques have been proposed, which are based on augmenting the training data by creating synthetic images via mixing or/and cutting samples from different classes.  In MixUp~\cite{zhang2018mixup} the input images and their target labels are interpolated using the same randomly chosen factor. \cite{Verma2018ManifoldMB} extends~\cite{zhang2018mixup} by performing interpolation not only in the input layer but also in the intermediate layers. CutOut~\cite{Devries2017ImprovedRO} augments an image by masking a rectangular region to zero. Differently, CutMix \cite{Yun2019CutMixRS} 
augments training data by creating synthetic images via cutting and pasting patches from image samples of different classes, marrying the best aspects of MixUp and CutOut. 
Other works employ the Mix\&Cut approaches for consistency regularization~\cite{Verma2019InterpolationCT,Berthelot2019MixMatchAH,Zhang2019ConsistencyRF}, 
i.e. penalizing the classification network sensitivity to samples generated via MixUp or CutOut~\cite{zhang2018mixup,Devries2017ImprovedRO}. In our work, we %
propose the consistency regularization under the CutMix transformation in the pixel output space of our U-Net discriminator. This helps to improve its localization quality and induce it to attend to non-discriminative differences between real and fake regions. %

\section{U-Net GAN Model}\label{method}

A "vanilla" GAN consists of two networks: a generator $G$ and a discriminator $D$,
trained by minimizing the following competing objectives in an alternating manner:
\begin{align}
\begin{array}{ll}
\mathcal{L}_{D} = -\mathbb{E}_x[\log D(x)]- \mathbb{E}_z[\log (1-D(G(z)))],  \\
\mathcal{L}_{G} = -\mathbb{E}_z[\log D(G(z))]\footnotemark.
\end{array} \label{eq:loss_d_g}
\end{align}
\footnotetext{This formulation is originally proposed as non-saturating (NS) GAN in \cite{goodfellow2014generative}.}
$G$ aims to map a latent variable $z \sim p(z)$ sampled from a prior distribution to a realistic-looking image, while $D$ aims to distinguish between real $x$ and generated $G(z)$ images. Ordinarily, $G$ and $D$ are modeled as a decoder and an encoder convolutional network, respectively.

While there are many variations of the GAN objective function and its network architectures \cite{Kurach2018GANlandscape,LucicEqualGANs},
in this paper we focus on improving the discriminator network. In Section~\ref{subsec:method-unet}, we propose to alter the $D$ architecture from a standard classification network to an encoder-decoder network -- U-Net \cite{Ronneberger2015UNetCN}, leaving the underlying basic architecture of $D$ -- the encoder part -- untouched. The proposed discriminator allows to maintain both global and local data representation, providing more informative feedback to the generator.
Empowered by local per-pixel feedback of the U-Net decoder module, in Section~\ref{subsec:method-cutmix} we further propose a consistency regularization technique, penalizing per-pixel inconsistent predictions of the discriminator under the CutMix transformations~\cite{Yun2019CutMixRS} of real and fake images.
This helps to improve the localization quality of the U-Net discriminator and induce it to attend more to semantic and structural changes between real and fake samples. We call our model \emph{U-Net GAN}.
Note that our method is compatible with most GAN models as it does not modify the generator in any way and leaves the original GAN objective intact.

\begin{figure}[t]
\begin{centering}
\setlength{\tabcolsep}{0.1em}
\renewcommand{\arraystretch}{0}
\par\end{centering}
\begin{centering}
\vspace{-1em}
\hfill{}%

\includegraphics[width=1\columnwidth]{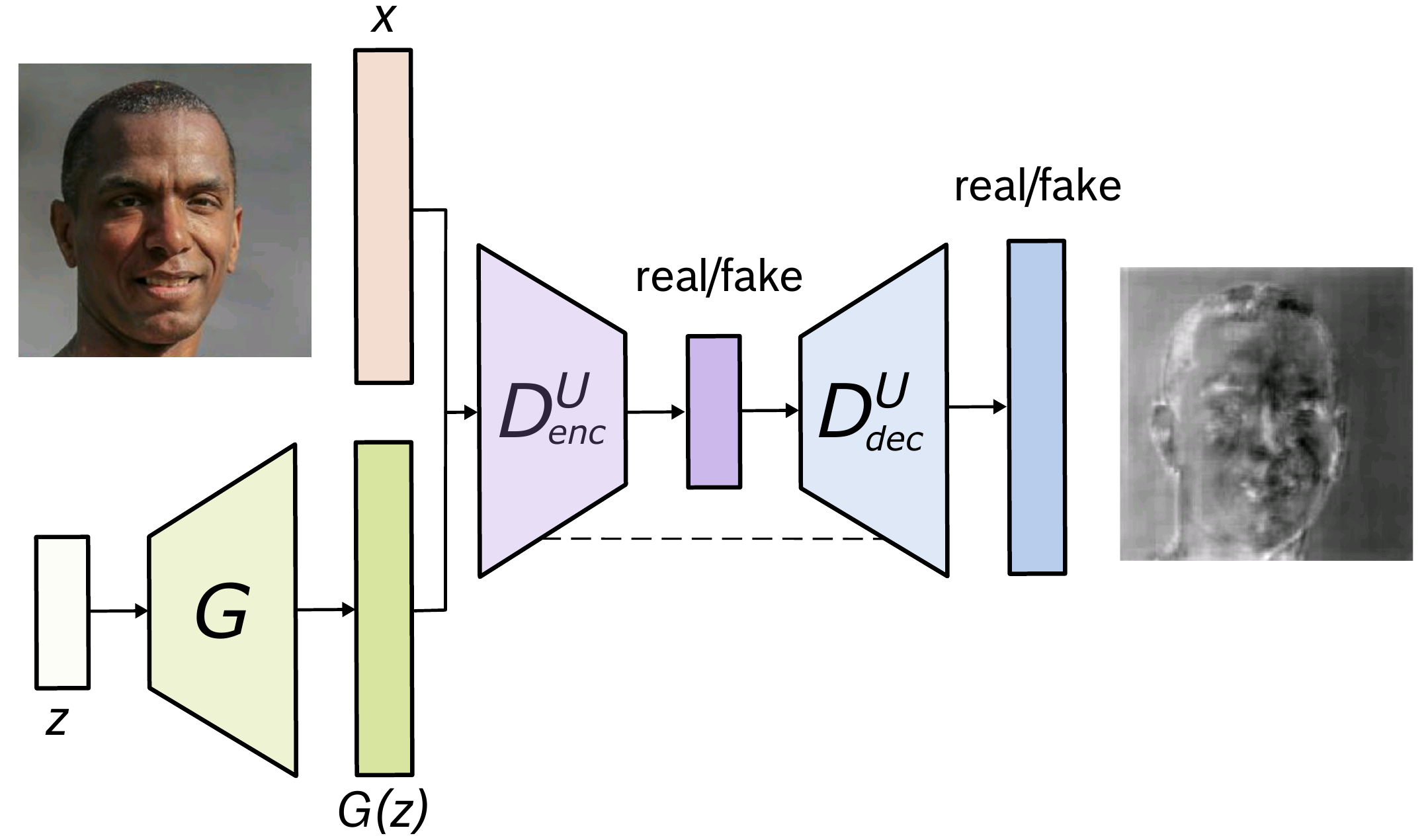}
\par\end{centering}
\vspace{-0.5em}
\caption{U-Net GAN. The proposed U-Net discriminator classifies the input images on a global and local \emph{per-pixel} level. Due to the skip-connections between the encoder and the decoder (dashed line), the channels in the output layer contain both high- and low-level information. Brighter colors in the decoder output correspond to the discriminator confidence of pixel being real (and darker of being fake).}

\label{fig:method_overview}
\vspace{-0.5em}
\end{figure}

\subsection{U-Net Based Discriminator}\label{subsec:method-unet}

Encoder-decoder networks \cite{badrinarayanan2017segnet,Ronneberger2015UNetCN} constitute a powerful method for dense prediction. %
U-Nets \cite{Ronneberger2015UNetCN} in particular have demonstrated state-of-art performance in many complex image segmentation tasks. In these methods, similarly to image classification networks, the encoder progressively downsamples the input, capturing the global image context.
The decoder performs progressive upsampling, matching the output resolution to the input one and thus enabling precise localization. Skip connections route data between the matching resolutions of the two modules, improving further the ability of the network to accurately segment fine details.

Analogously, in this work, we propose to extend a discriminator to form a U-Net, by reusing building blocks of the original discriminator classification network as an encoder part and building blocks of the generator network as the decoder part. In other words, the discriminator now consists of the original downsampling network and a new upsampling network. The two modules are connected via a bottleneck, as well as skip-connections that copy and concatenate feature maps from the encoder and the decoder modules, following \cite{Ronneberger2015UNetCN}.  We will refer to this discriminator as $D^{U}$. While the original $D(x)$ classifies the input image $x$ into being real and fake, the U-Net discriminator $D^{U}(x)$ additionally performs this classification on a \textit{per-pixel} basis, segmenting image $x$ into real and fake regions, along with the original image classification of $x$ from the encoder, see Figure \ref{fig:method_overview}. This enables the discriminator to learn both global and local differences between real and fake images.

Hereafter, we refer to the original encoder module of the discriminator as $D^{U}_{enc}$ and to the introduced decoder module as $D^{U}_{dec}$. The new discriminator loss is now can be computed by taking the decisions from both $D^{U}_{enc}$ and $D^{U}_{dec}$:
\begin{align}
& \mathcal{L}_{D^U}  = \mathcal{L}_{D^U_{enc}} + \mathcal{L}_{D^U_{dec}}, \label{eq:loss_d_u}
\end{align}
where similarly to Eq.~\ref{eq:loss_d_g} the loss for the encoder ${L}_{D^U_{enc}}$ is computed from the scalar output of $D^U_{enc}$:
{\medmuskip=0mu
\thinmuskip=0mu
\thickmuskip=0mu
\begin{align}
&\hspace{-0.5em}\mathcal{L}_{D^U_{enc}}= -\mathbb{E}_x[\log D^U_{enc}(x)]- \mathbb{E}_z[\log (1-D^U_{enc}(G(z)))],\label{eq:loss_d_u_enc}
\end{align} }and the loss for the decoder ${L}_{D^U_{enc}}$ is computed as the mean decision over all pixels:
\begin{multline}
\mathcal{L}_{D^U_{dec}} = -\mathbb{E}_x\Big[\sum_{i,j}\log [D^U_{dec}(x)]_{i,j}\Big] \\
- \mathbb{E}_z\Big[\sum_{i,j}\log (1-[D^U_{dec}(G(z))]_{i,j})\Big]. \label{eq:loss_d_u_dec}
\end{multline}
Here, $[D^U_{dec}(x)]_{i,j}$ and $[D^U_{dec}(G(z))]_{i,j}$ refer to the discriminator decision at pixel $(i,j)$.  These per-pixel outputs of $D^U_{dec}$ are derived based on global information from high-level features, enabled through the process of upsampling from the bottleneck, as well as more local information from low-level features, mediated by the skip connections from the intermediate layers of the encoder network.

Correspondingly, the generator objective becomes:
\begin{multline}
\mathcal{L}_{G} =  -\mathbb{E}_z\Big[\log D^U_{enc}(G(z))  \\
+ \sum_{i,j}\log [D^U_{dec}(G(z))]_{i,j}\Big],  \label{eq:loss_g_u}
\end{multline}
encouraging the generator to focus on both global structures and local details while synthesizing images in order to fool the more powerful discriminator $D^U$.

\subsection{Consistency Regularization} \label{subsec:method-cutmix}

\begin{figure}
		\setlength{\tabcolsep}{0em}
		\renewcommand{\arraystretch}{0.7}
		\vspace{-1em}

	\begin{centering}

\begin{tabular}{@{\hskip -0.2in}c@{ }c@{ }c@{ }c@{ }c@{}}

	 & \multicolumn{2}{c}{\footnotesize{} Real}  & \multicolumn{2}{c}{\footnotesize{} Fake} \\
	  \vspace{0.5em}
  \begin{minipage}{0.19\columnwidth} 	\hfill{} \begin{tabular}{c} {	\footnotesize{} Original} \\
{\footnotesize{} images}\end{tabular} 	\hfill{} \end{minipage}  &
 \begin{minipage}{0.2\columnwidth} 	\includegraphics[width=1\columnwidth, height=1\columnwidth]{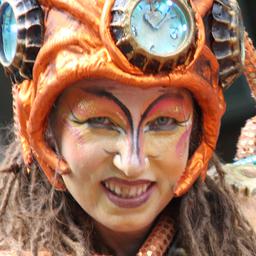}  \end{minipage} & 
  \begin{minipage}{0.2\columnwidth} 	\includegraphics[width=1\columnwidth, height=1\columnwidth]{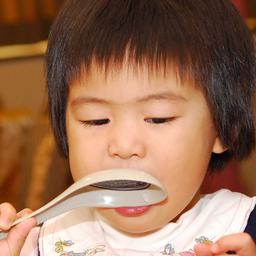}  \end{minipage} &
 \begin{minipage}{0.2\columnwidth} 	\includegraphics[width=1\columnwidth, height=1\columnwidth]{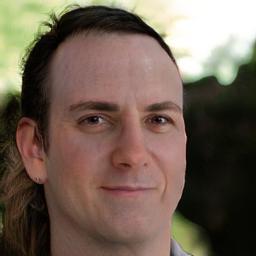}  \end{minipage} &
 \begin{minipage}{0.2\columnwidth} 	\includegraphics[width=1\columnwidth, height=1\columnwidth]{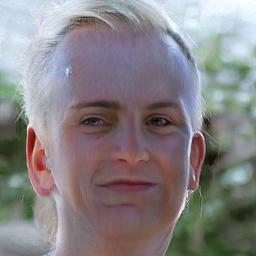}  \end{minipage}\\
 
\vspace{0.0em}
\begin{minipage}{0.19\columnwidth} 	\hfill{} \begin{tabular}{c}	\footnotesize{}Real/fake \\
		\footnotesize{} ratio $r$ 	\end{tabular} 	\hfill{} \end{minipage} & 0.28 & 0.68 &  0.31 &  0.51 \\

\vspace{0.5em}

\begin{minipage}{0.19\columnwidth}	\hfill{}	\begin{tabular}{c}	\footnotesize{} Mask \\
		\footnotesize{} $\mathrm{M}$ \end{tabular} 	\hfill{} \end{minipage} &
\begin{minipage}{0.2\columnwidth}	\includegraphics[width=1\columnwidth, height=1\columnwidth, cfbox=verylightgray 0.2pt 0.2pt]{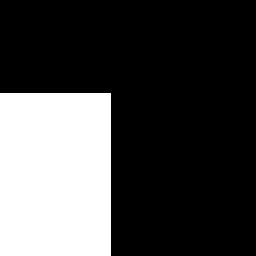} \end{minipage}& 
\begin{minipage}{0.2\columnwidth}	\includegraphics[width=1\columnwidth, height=1\columnwidth, cfbox=verylightgray 0.2pt 0.2pt]{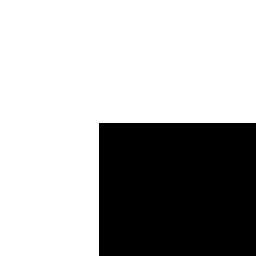} \end{minipage} & 
\begin{minipage}{0.2\columnwidth}	\includegraphics[width=1\columnwidth, height=1\columnwidth, cfbox=verylightgray 0.2pt 0.2pt]{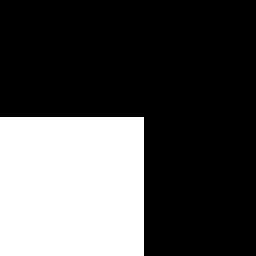} \end{minipage} & \begin{minipage}{0.2\columnwidth}	\includegraphics[width=1\columnwidth, height=1\columnwidth, cfbox=verylightgray 0.2pt 0.2pt]{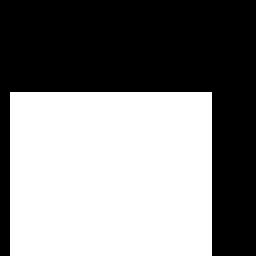} \end{minipage}\\

\vspace{0.5em}

  \begin{minipage}{0.19\columnwidth}	\hfill{}	\begin{tabular}{c}	\footnotesize{} CutMix \\
	\footnotesize{} images \end{tabular} 	\hfill{} \end{minipage} &
  \begin{minipage}{0.2\columnwidth}	\includegraphics[width=1\columnwidth, height=1\columnwidth]{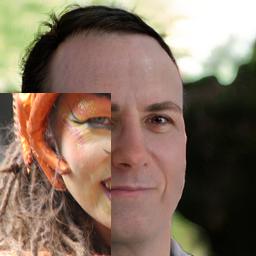} \end{minipage}& 
   \begin{minipage}{0.2\columnwidth}	\includegraphics[width=1\columnwidth, height=1\columnwidth]{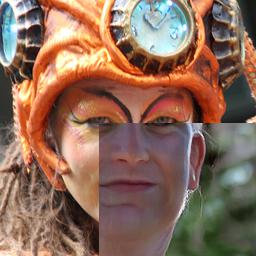} \end{minipage} & 
  \begin{minipage}{0.2\columnwidth}	\includegraphics[width=1\columnwidth, height=1\columnwidth]{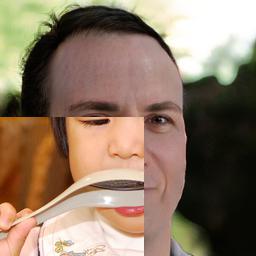} \end{minipage} & 
  \begin{minipage}{0.2\columnwidth}	\includegraphics[width=1\columnwidth, height=1\columnwidth]{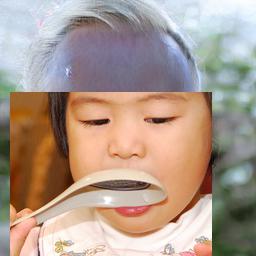} \end{minipage}\\
 \vspace{0.1em}

   \begin{minipage}{0.19\columnwidth} 	\hfill{} \begin{tabular}{c}	\footnotesize{} $D^U_{dec}$ segm. \\
	\footnotesize{} map 	\end{tabular} 	\hfill{} \end{minipage} &
\begin{minipage}{0.2\columnwidth} \includegraphics[width=1\columnwidth, height=1\columnwidth]{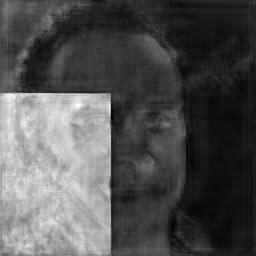} \end{minipage} &
\begin{minipage}{0.2\columnwidth} \includegraphics[width=1\columnwidth, height=1\columnwidth]{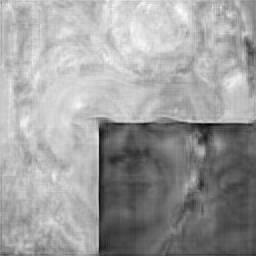} \end{minipage} &
\begin{minipage}{0.2\columnwidth} \includegraphics[width=1\columnwidth, height=1\columnwidth]{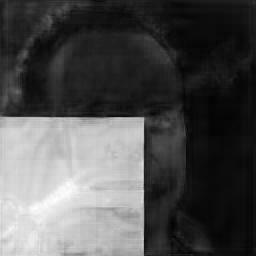}  \end{minipage} &
  \begin{minipage}{0.2\columnwidth} \includegraphics[width=1\columnwidth, height=1\columnwidth]{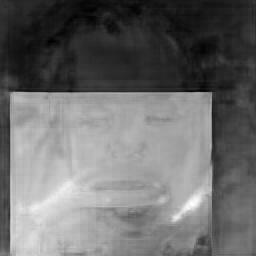} \end{minipage}   \\

   \begin{minipage}{0.19\columnwidth} 	\hfill{} \begin{tabular}{c}	\footnotesize{} $D^U_{enc}$ class. \\
	\footnotesize{} score 	\end{tabular} 	\hfill{} \end{minipage} & 0.31 & 0.60 &  0.36 &  0.43 \\

		\end{tabular}

	\par\end{centering}

	\vspace{-0.5em}

	\caption{\label{fig:cutmix} Visualization of the CutMix augmentation and the predictions of the U-Net discriminator on CutMix images. 1st row: real and fake samples. 2nd\&3rd rows: sampled real/fake CutMix ratio $r$ and corresponding binary masks $\mathrm{M}$ (color code: white for real, black for fake). 4th row: generated CutMix images from real and fake samples. 5th\&6th row: the corresponding real/fake segmentation maps of $D^U$ with its predicted classification scores. }
	\vspace{-0.5em}
\end{figure}

Here we present the consistency regularization technique for the U-Net based discriminator introduced in the previous section. %
The per-pixel decision of the well-trained $D^U$ discriminator should be equivariant under any class-domain-altering transformations of images. However, this property is not explicitly guaranteed.
To enable it, the discriminator should be regularized to focus more on semantic and structural changes between real and fake samples and to pay less attention to arbitrary class-domain-preserving perturbations.
Therefore, we propose
the consistency regularization of the $D^U$
discriminator, explicitly encouraging the decoder module $D^U_{dec}$ to output equivariant predictions under the
CutMix transformations~\cite{Yun2019CutMixRS} of real and fake samples.
The CutMix augmentation creates synthetic images via cutting and pasting patches from images of different classes.
We choose CutMix among other Mix\&Cut strategies (cf. Section \ref{related_work}) as it does not alter the real and fake image patches used for mixing, in contrast to~\cite{zhang2018mixup}, preserving their original class domain, and provides a large variety of possible outputs. We visualize the CutMix augmentation strategy and the $D^U$ predictions in Figure \ref{fig:cutmix}.

Following~\cite{Yun2019CutMixRS}, we
synthesize a new training sample $\tilde{x}$ for the discriminator $D^U$ by mixing $x$ and $G(z) \in \mathbb{R}^{W\times H \times C}$ with the mask $\mathrm{M}$:
\begin{align}
\begin{array}{ll}
& \tilde{x} = \mathrm{mix} (x, G(z) , \mathrm{M}), \\
& \mathrm{mix} (x, G(z) , \mathrm{M}) = \mathrm{M}  \odot x + (\mathrm{1}-\mathrm{M}) \odot G(z), \end{array} \label{eq:cutmix}
\end{align}
where $\mathrm{M} \in \{0,1\}^{W\times H}$ is the binary mask indicating if the pixel $(i,j)$ comes from the real ($\mathrm{M}_{i,j}=1$) or fake ($\mathrm{M}_{i,j}=0$) image, $\mathrm{1}$ is a binary mask filled with ones, and $\odot$ is an element-wise multiplication.
In contrast to \cite{Yun2019CutMixRS}, the class label $c \in \{0,1\}$ for the new CutMix image $\tilde{x}$ is set to be fake, i.e. $c=0$. Globally the mixed synthetic image should be recognized as fake by the encoder $D^U_{enc}$, otherwise the generator can learn to introduce the CutMix augmentation
into generated samples, causing undesirable artifacts.
Note that for the synthetic sample $\tilde{x}$, $c=0$ and $\mathrm{M}$ are the ground truth for the encoder and decoder modules of the discriminator $D^U$, respectively.

Given the CutMix operation in Eq.~\ref{eq:cutmix}, we train the discriminator to provide consistent per-pixel predictions, i.e.
{\thickmuskip=2mu $D^U_{dec}\big(\mathrm{mix}(x, G(z), \mathrm{M})\big) \approx \mathrm{mix}\big(D^U_{dec}(x),D^U_{dec}(G(z)), \mathrm{M}\big)$}, by introducing the consistency regularization loss term in the discriminator objective:
\begin{multline}
\mathcal{L}^{cons}_{D^U_{dec}} =  \Big\|D^U_{dec}\Big(\mathrm{mix} (x, G(z), \mathrm{M})\Big) \\
- \mathrm{mix} \Big(D^U_{dec}(x), D^U_{dec}(G(z)), \mathrm{M}\Big)\Big\|^2,  \label{eq:loss_cr}
\end{multline}
where denotes $\|\cdot\|$ the $L^2$ norm.
This consistency loss is then taken between the per-pixel output of $D^U_{dec}$ on the
CutMix image and the CutMix between outputs of the $D^U_{dec}$ on real and fake images, penalizing the discriminator for inconsistent predictions.

We add the loss term in Eq.~\ref{eq:loss_cr} to the discriminator objective in Eq.~\ref{eq:loss_d_u} with a weighting hyper-parameter $\lambda$:
\begin{align}
& \mathcal{L}_{D^U}  = \mathcal{L}_{D^U_{enc}} + \mathcal{L}_{D^U_{dec}} +\lambda \mathcal{L}^{cons}_{D^U_{dec}}. \label{eq:fin_loss_d_u}.
\end{align}
The generator objective $\mathcal{L}_{G}$ remains unchanged, see Eq.~\ref{eq:loss_g_u}.

In addition to the proposed consistency regularization, we also use CutMix samples for training both the encoder and decoder modules of $D^U$.  Note that for the U-Net GAN we use the non-saturating GAN objective formulation~\cite{goodfellow2014generative}. However, the introduced consistency regularization as well as the U-Net architecture of the discriminator can be combined with any other adversarial losses of the generator and discriminator~\cite{Arjovsky2017WGAN,Lim2017GeometricG,Nowozin2016fGANTG}.

\begin{figure*}
\begin{centering}
\setlength{\tabcolsep}{0.1em}
\renewcommand{\arraystretch}{0}
\par\end{centering}
\begin{centering}
\vspace{-1.5em}
\hfill{}%
\begin{tabular}{c@{\hskip 0.05in}c@{\hskip 0.05in}c@{\hskip 0.05in}c@{\hskip 0.05in}c@{\hskip 0.05in}c}
\includegraphics[width=0.85\textwidth]{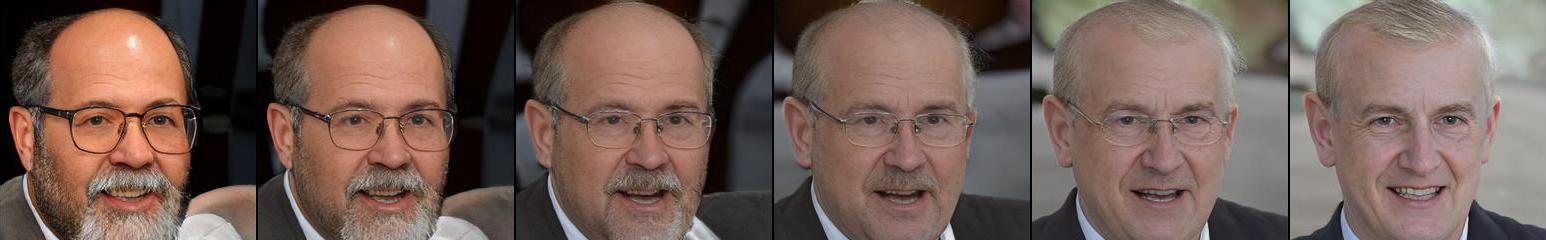}\tabularnewline
\includegraphics[width=0.85\textwidth]{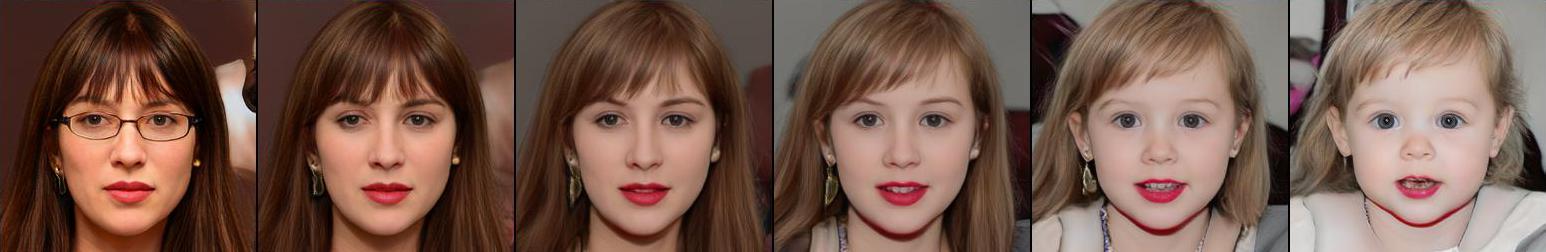}\tabularnewline
\includegraphics[width=0.85\textwidth]{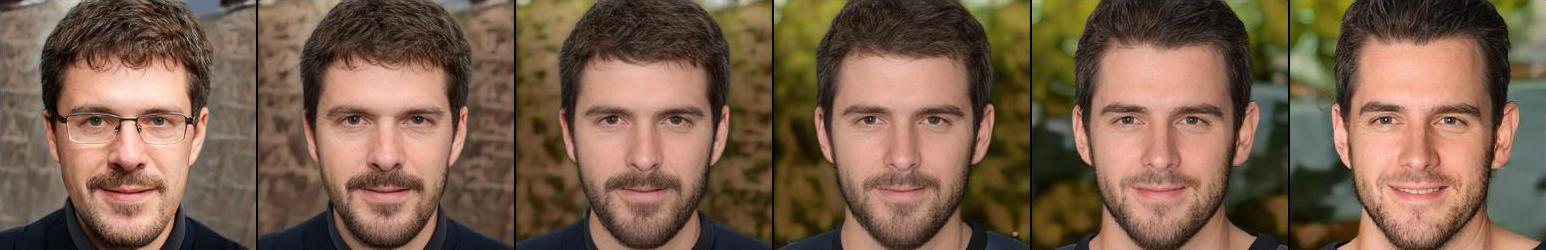}\tabularnewline
\includegraphics[width=0.85\textwidth]{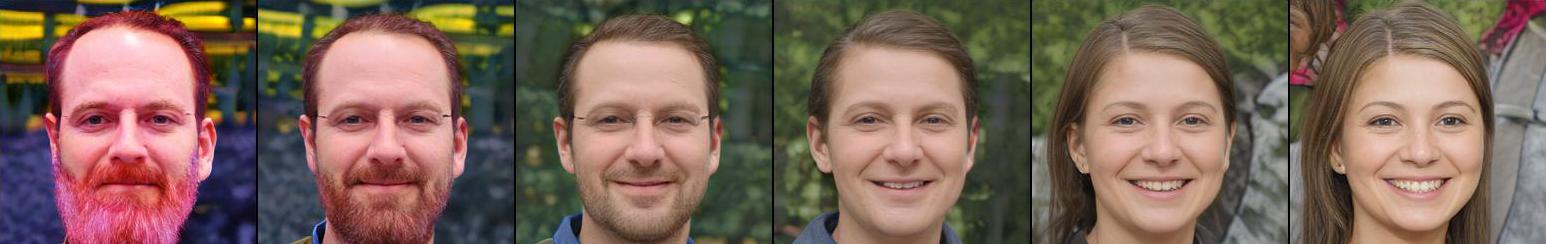}\tabularnewline
\includegraphics[width=0.85\textwidth]{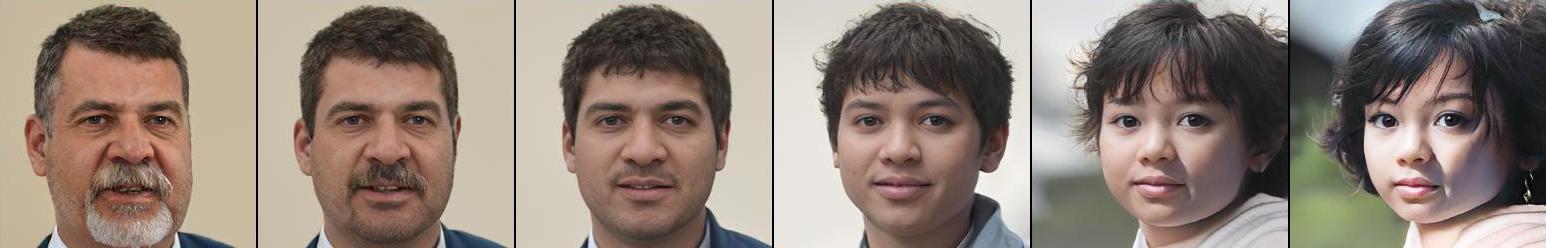}\tabularnewline
\includegraphics[width=0.85\textwidth]{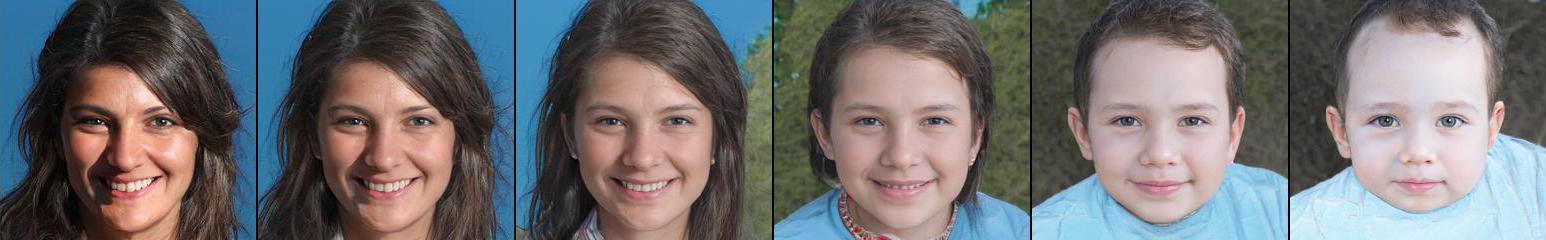}\tabularnewline
\includegraphics[width=0.85\textwidth]{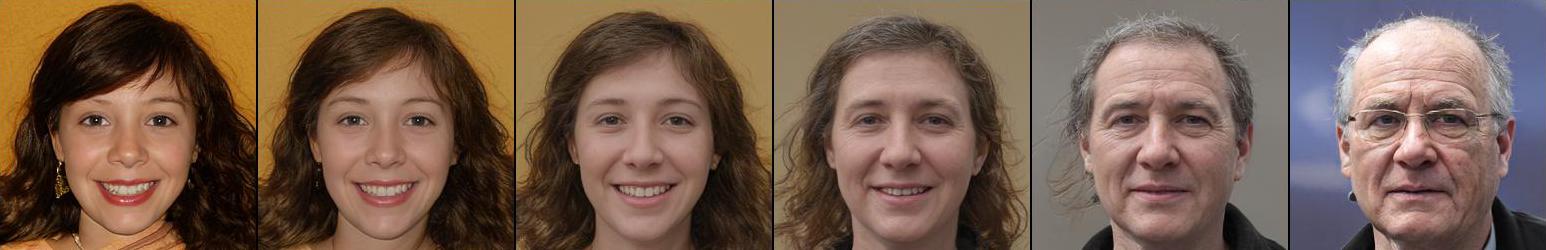}\tabularnewline
\includegraphics[width=0.85\textwidth]{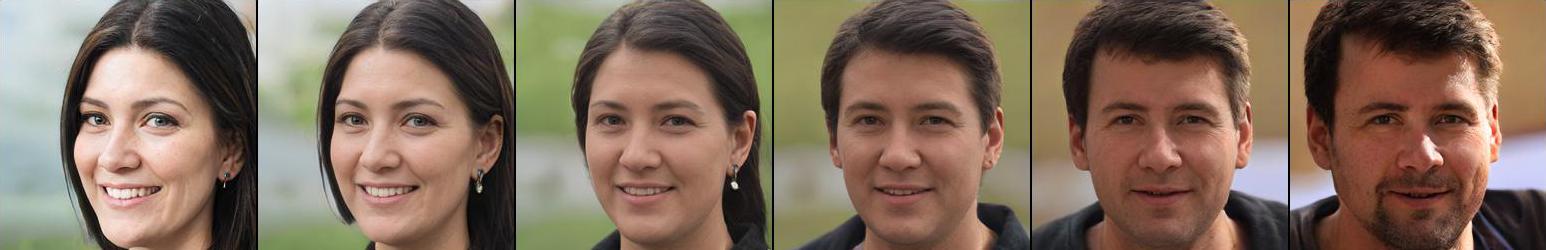}\tabularnewline
\includegraphics[width=0.85\textwidth]{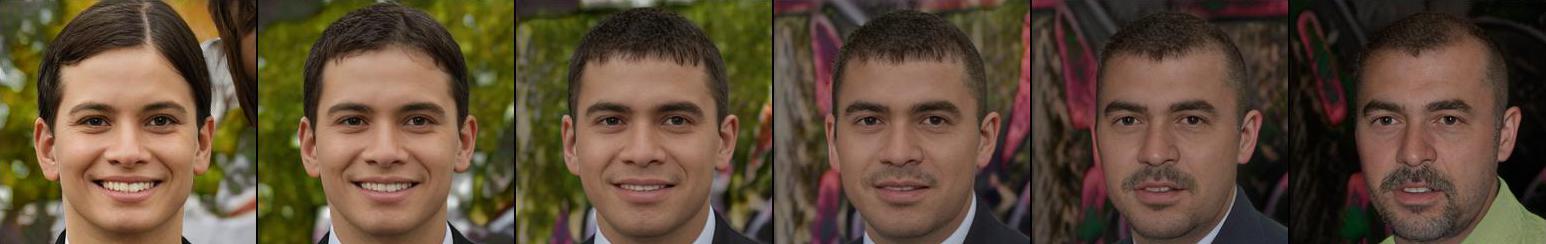}\tabularnewline

\end{tabular}\hfill{}
\par\end{centering}
\vspace{-0.5em}
\caption{\label{fig:qual_results} Images generated with U-Net GAN trained on FFHQ with resolution $256 \times 256$ when interpolating in the latent space between two synthetic samples (left to right). Note the high quality of synthetic samples and very smooth interpolations, maintaining \textit{global} and \textit{local} realism.} %
\vspace{0em}
\end{figure*}

\subsection{Implementation}
Here we discuss implementation details of the U-Net GAN model proposed in Section~\ref{subsec:method-unet} and~\ref{subsec:method-cutmix}.

\paragraph{U-Net based discriminator}
We build upon the recent state-of-the-art BigGAN model~\cite{Brock2019}, and extend its discriminator with our proposed changes.
We adopt the BigGAN generator and discriminator architectures for the $256\times 256$ (and $128\times 128$) resolution with a channel multiplier $ch=64$, as described in detail in \cite{Brock2019}. The original BigGAN discriminator downsamples the input image to a feature map of dimensions $16ch\times 4\times 4$, on which global sum pooling is applied to derive a $16ch$ dimensional feature vector that is classified into real or fake. In order to turn the discriminator into a U-Net, we copy the generator architecture and append it to the $4\times4$ output of the discriminator.
In effect, the features are successively upsampled via ResNet blocks until the original image resolution ($H\times W$) is reached. %
To make the U-Net complete, the input to every decoder ResNet block is concatenated with the output features of the encoder blocks that share the same intermediate resolution. In this way, high-level and low-level information are effectively integrated on the way to the output feature map.
Hereby, the decoder architecture is almost identical to the generator, with the exception of that we change the number of channels of the final output from $3$ to $ch$, append a final block of $1\times1$ convolutions to produce the $1\times H\times W$ output map, and do not use class-conditional BatchNorm~\cite{Vries2017ModulatingEV,Dumoulin2016ALR} in the decoder, nor the encoder.
Similarly to \cite{Brock2019}, we provide class information to $D^U$ with projection \cite{miyato2018cgans} to the $ch$-dimensional channel features
of the U-Net encoder and decoder output.
In contrast to \cite{Brock2019} and in alignment with \cite{Chen2018OnSM}, we find it beneficial not to use a hierarchical latent space, but to directly feed the same input vector $z$ to BatchNorm at every layer in the generator.
Lastly, we also remove the self-attention layer in both encoder and decoder, as in our experiments they did not contribute to the performance but led to memory overhead.
While the original BigGAN is a class-conditional model, we additionally devise an unconditional version for our experiments. For the unconditional model, we replace class-conditional BatchNorm with self-modulation \cite{Chen2018OnSM}, where the BatchNorm parameters are conditioned only on the latent vector $z$, and do not use the class projection of \cite{miyato2018cgans} in the discriminator.

All these modifications leave us with a two-headed discriminator.
While the decoder head is already sufficient to train the network, we find it beneficial to compute the GAN loss at both heads with equal weight. Analogously to BigGAN, we keep the hinge loss \cite{Zhang_SAGAN19} in all basic U-Net models, while the models that also employ the consistency regularization in the decoder output space benefit from using the non-saturating loss~\cite{goodfellow2014generative}.
Our implementation builds on top of the original BigGAN PyTorch implementation\footnote{\url{https://github.com/ajbrock/BigGAN-PyTorch}}.

\begin{figure*}[h]
\begin{centering}
\setlength{\tabcolsep}{0.0em}
\renewcommand{\arraystretch}{0}
\par\end{centering}
\begin{centering}
\vspace{-1em}
\hfill{}%
\begin{tabular}{@{}c@{\hskip 0.05in}c@{\hskip 0.05in}c@{\hskip 0.05in}c@{\hskip 0.05in}c@{\hskip 0.05in}c@{\hskip 0.05in}c@{}}
\includegraphics[width=0.13\textwidth]{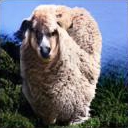} &
\includegraphics[width=0.13\textwidth]{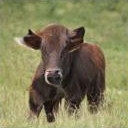} &
\includegraphics[width=0.13\textwidth]{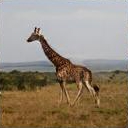} &
\includegraphics[width=0.13\textwidth]{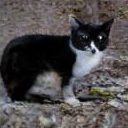} &
\includegraphics[width=0.13\textwidth]{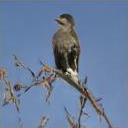}  & %
\includegraphics[width=0.13\textwidth]{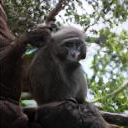}  &
\includegraphics[width=0.13\textwidth]{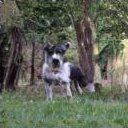} 

\tabularnewline

\includegraphics[width=0.13\textwidth]{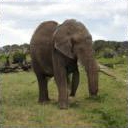} &
\includegraphics[width=0.13\textwidth]{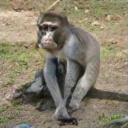} &
\includegraphics[width=0.13\textwidth]{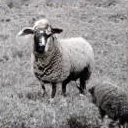} &
\includegraphics[width=0.13\textwidth]{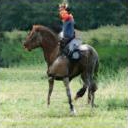}  &
\includegraphics[width=0.13\textwidth]{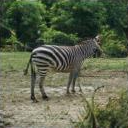}  &
\includegraphics[width=0.13\textwidth]{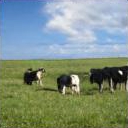}  &
\includegraphics[width=0.13\textwidth]{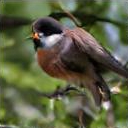}
\tabularnewline

\end{tabular}\hfill{}
\par\end{centering}
\vspace{-0.5em}
\caption{\label{fig:coco_pics} Images generated with U-Net GAN trained on COCO-Animals with resolution $128 \times 128$.
 }
\vspace{-0.5em}
\end{figure*}

\paragraph{Consistency regularization}

For each training iteration a mini-batch of CutMix images $(\tilde{x}, c=0, \mathrm{M})$ is created with probability $p_{mix}$. This probability is increased linearly from $0$ to $0.5$ between the first $n$ epochs in order to give the generator time to learn how to synthesize more real looking samples and not to give the discriminator too much power from the start. CutMix images are created from the existing real and fake images in the mini-batch using binary masks $\mathrm{M}$.
For sampling $\mathrm{M}$, we use the original CutMix implementation\footnote{\url{https://github.com/clovaai/CutMix-PyTorch}}: first sampling the combination ratio $r$ between the real and generated images from the uniform distribution $(0,1)$ and then uniformly sample the bounding box coordinates for the cropping regions of $x$ and $G(z)$ to preserve the $r$ ratio, i.e. $r = \frac{|\mathrm{M}|}{W*H}$ (see Figure \ref{fig:cutmix}). Binary masks $\mathrm{M}$ also denote the target for the decoder $D^U_{dec}$, while we use \textit{fake}, i.e $c=0$, as the target for the encoder $D^U_{enc}$. We set $\lambda=1.0$ as it showed empirically to be a good choice. Note that the consistency regularization does not impose much overhead during training. Extra computational cost comes only from feeding additional CutMix images through the discriminator while updating its parameters.

\section{Experiments}\label{experiments}

\subsection{Experimental Setup}\label{subsec:exp_setup}

\paragraph{Datasets}
We consider three datasets: FFHQ \cite{Karras2018ASG}, CelebA~\cite{Liu_Celeba} and the subset of the COCO~\cite{Lin2014MicrosoftCC} and OpenImages \cite{OpenImages} images containing  animal classes, which we will further on refer to as COCO-Animals. We use FFHQ and CelebA for unconditional image synthesis and COCO-Animals for class-conditional image synthesis, where the class label is used. We experiment with $256\times 256$ resolution for FFHQ and $128\times 128$ for CelebA and COCO-Animals.

CelebA is a human face dataset of $200$\unit{k} images, featuring $\sim10$\unit{k} different celebrities with a variety of facial poses and expressions. Similarly, FFHQ is a more recent dataset of human faces, consisting of $70$\unit{k} high-quality images with higher variation in terms of age, ethnicity, accessories, and viewpoints. 
The proposed COCO-Animals dataset consists of $\sim 38$\unit{k} training images belonging to $10$ animal classes, where 
we choose COCO and OpenImages (using the human verified subset with mask annotations) samples in the categories \textit{bird, cat, dog, horse, cow, sheep, giraffe, zebra, elephant}, and \textit{monkey}.  With its relatively small size and imbalanced number of images per class as well as due to its variation in poses, shapes, number of objects, and backgrounds, COCO-Animals presents a challenging task for class-conditional image synthesis.
We choose to create this dataset in order to perform conditional image generation in the mid- to high-resolution regime, with a reasonable computational budget and feasible training time. Other datasets in this order of size either have too few examples per class (e.g. AwA~\cite{Xian2017ZeroShotLC}) or too little inter- and intra-class variability. In contrast, the intra-class variability of COCO-Animals is very high for certain classes, e.g. bird and monkey, which span many subspecies. For more details, we refer to Section \ref{sec:syn samples_coco} in the supplementary material.

\begin{table}
	\vspace{-1em}
	\setlength{\tabcolsep}{0.0em}
	\renewcommand{\arraystretch}{1.1}
	\centering

\hskip -0.15in		\begin{tabular}{lcc|cc||cc|cc}
		\multirow{3}{*}{\normalsize{} Method } & \multicolumn{4}{c}{\normalsize{} FFHQ}	& \multicolumn{4}{c}{\normalsize{}{COCO-Animals}} \tabularnewline

		& \multicolumn{2}{c|}{\normalsize{} Best } & \multicolumn{2}{c||}{\normalsize{} Median }  & \multicolumn{2}{c|}{\normalsize{} Best} & \multicolumn{2}{c}{\normalsize{} Median}\tabularnewline
	\small{}   & \small{} FID$\downarrow$ & \small{} IS$\uparrow$ 	& \small{} FID$\downarrow$ & \small{} IS$\uparrow$  & \small{} FID$\downarrow$  & \small{} IS$\uparrow$  & \small{} FID$\downarrow$ & \small{} IS$\uparrow$ \tabularnewline
	
		\hline 	\hline
		\normalsize{} BigGAN \cite{Brock2019} & \normalsize{} 11.48 & \normalsize{} 3.97 & \normalsize{} 12.42 & \normalsize{} 4.02  & \normalsize{} 16.37  & \normalsize{} 11.77  & \normalsize{} 16.55  & \normalsize{} 11.78 \tabularnewline
		\normalsize{} U-Net GAN & \normalsize{} \textbf{7.48} & \normalsize{} \textbf{4.46}	& \normalsize{}  \textbf{7.63} & \normalsize{}  \textbf{4.47} &  \normalsize{} \textbf{13.73} &  \normalsize{} \textbf{12.29} & \normalsize{} \textbf{13.87}  & \normalsize{} \textbf{12.31} \tabularnewline
	\end{tabular}%
	
	\vspace{-0.5em}
    \caption{Evaluation results on FFHQ and COCO-Animals. We report the best and median FID score across 5 runs and its corresponding IS, see Section~\ref{subsec:exp_results} for discussion.
    } \label{table:fid_overview} %
	\vspace{-0.0em}
\end{table}
\begin{figure}
\begin{centering}
\setlength{\tabcolsep}{0em}
\renewcommand{\arraystretch}{0}
\par\end{centering}
\begin{centering}
\vspace{-1em}
\hfill{}%
\begin{tabular}{@{}c@{}c@{}}
	FFHQ & COCO-Animals \\
\includegraphics[width=0.49\columnwidth]{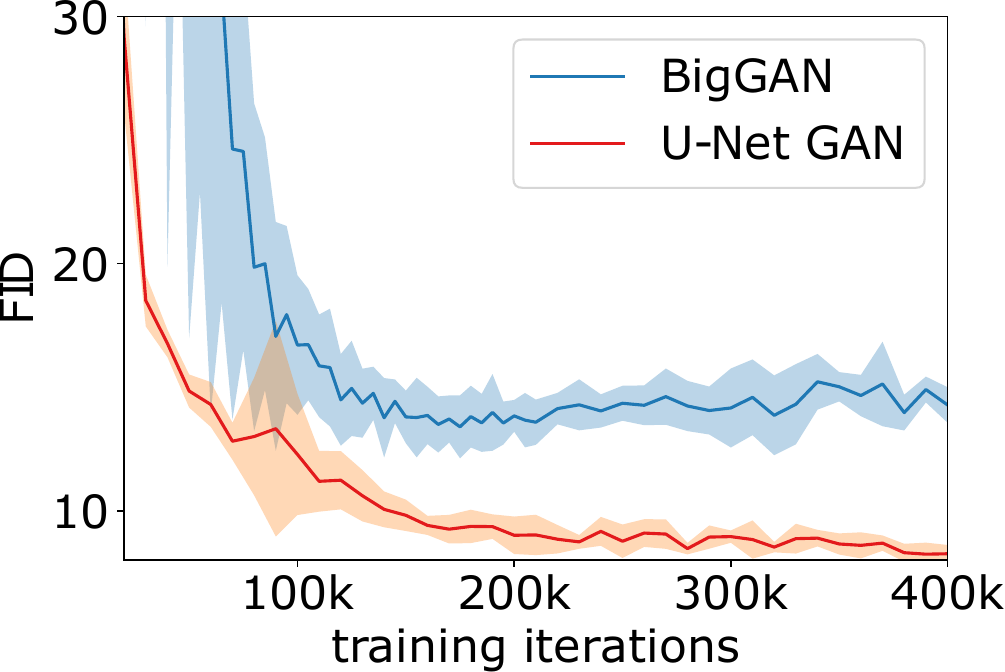}&
\includegraphics[width=0.49\columnwidth]{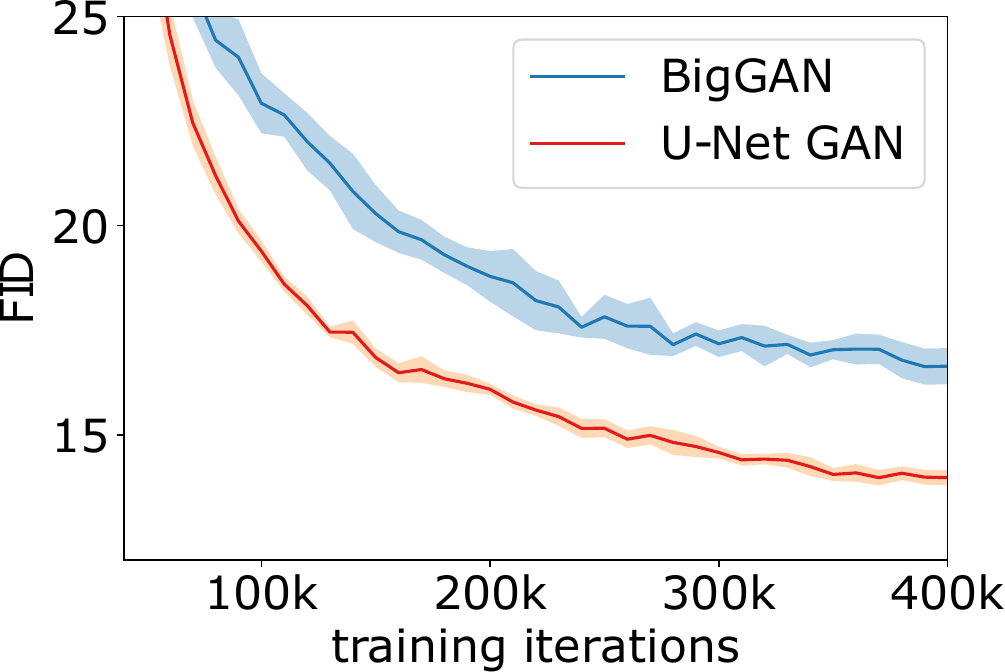}
\end{tabular}\hfill{}
\par\end{centering}
\vspace{-0.5em}
\caption{\label{fig:fid_curves} FID curves over iterations of the BigGAN model (blue) and the proposed U-Net GAN (red). Depicted are the FID mean and standard deviation across 5 runs per setting.}	
\vspace{-1em}
\end{figure}

\paragraph{Evaluation metrics}
For quantitative evaluation we use the Fr{\'e}chet Inception distance (FID)~\cite{heuselttur2017} as the main metric, and additionally consider the Inception score (IS)~\cite{SalimansNeurIPS2016}. %
Between the two, FID is a more comprehensive metric, which has been shown to be more consistent
with human evaluation in assessing the realism and variation of the generated images~\cite{heuselttur2017}, while IS is limited by what the Inception classifier can recognise, which is directly linked to its training data \cite{Barratt2018ANO}. If one learns to generate something not present in the classifier’s training data (e.g. human faces) then IS can still be low despite generating high quality images since that image does not get classified as a distinct class.

In all our experiments, FID and IS are computed using $50$\unit{k} synthetic images, following~\cite{karras2018progressive}.
By default all reported numbers correspond to the best or median FID of five independent runs achieved with $400$\unit{k} training iterations for FFHQ and COCO-Animals, and $800$\unit{k} training iterations for CelebA. 
For evaluation, we employ moving averages of the generator weights following \cite{Brock2019,karras2018progressive}, with a decay of $0.9999$. Note that we do not use any truncation tricks or rejection sampling for image generation.

\paragraph{Training details}
We adopt the original training parameters of \cite{Brock2019}. In particular, we use a uniformly distributed noise vector $z\in[-1,1]^{140}$ as input to the generator, and the Adam optimizer~\cite{adamopt} with learning rates of $1e\text{-}4$ and $5e\text{-}4$ for $G$ and $D^{U}$. %
The number of warmup epochs $n$ for consistency regularization is chosen to be $200$ for COCO-Animals and $20$ for FFHQ and CelebA. 
In contrast to \cite{Brock2019}, we operate with considerably smaller mini-batch sizes: $20$ for FFHQ, $50$ for CelebA and $80$ for COCO-Animals. See Section \ref{sec:networks} in the supplementary material for more details.

\subsection{Results}\label{subsec:exp_results}

\begin{table}
	\vspace{-1em}
	\setlength{\tabcolsep}{0.0em}
	\renewcommand{\arraystretch}{1.1}
	\centering

	\begin{tabular}{lc@{ \hskip 0.05in}c}
		\normalsize{} Method & \normalsize{} COCO-Animals  & \normalsize{} FFHQ  \tabularnewline
		\hline 	\hline
		\normalsize{} BigGAN \cite{Brock2019} & \normalsize{} $16.55$  & \normalsize{} $12.42$ \tabularnewline
		\hline
		\normalsize{} U-Net based discriminator        & \normalsize{} $15.86$  & \normalsize{} $10.86$	\tabularnewline
		\normalsize{} \quad + CutMix augmentation      & \normalsize{} $14.95$  & \normalsize{} $10.30$ \tabularnewline
		\normalsize{} \quad\quad+ Consistency regularization & \normalsize{} $\mathbf{13.87}$  & \normalsize{}  $\mathbf{7.63}$ \tabularnewline
	\end{tabular}

\vspace{-0.5em}
    \caption{Ablation study of the U-Net GAN model on FFHQ and COCO-Animals. Shown are the median FID scores.
   The proposed components lead to better performance, on average improving the median FID by $3.7$ points over BigGAN~\cite{Brock2019}. See Section~\ref{subsec:exp_results} for discussion. 
}
   \label{table:fid_ablation} %
	\vspace{-1em}
\end{table}

We first test our proposed U-Net discriminator in two settings: unconditional image synthesis on FFHQ and 
class-conditional image synthesis on COCO-Animals, using the BigGAN model~\cite{Brock2019} as a baseline for comparison. We report our key results in Table \ref{table:fid_overview} and Figure~\ref{fig:fid_curves}. 

In the unconditional case, our model achieves the FID score of $7.48$, which is an improvement of $4.0$ FID points over the canonical BigGAN discriminator (see Table \ref{table:fid_overview}). In addition, the new U-Net discriminator also improves over the baseline in terms of the IS metric ($3.97$ vs. $4.46$).  
The same effect is observed for the conditional image generation setting. Here, our U-Net GAN achieves an FID of $13.73$, improving $2.64$ points over BigGAN, as well as increases the IS score from $11.77$ to $12.29$. 
Figure \ref{fig:fid_curves} visualizes the mean FID behaviour over the training across 5 independent runs.
From Figure \ref{fig:fid_curves} it is evident that the FID score drops for both models at the similar rate, with a constant offset for the U-Net GAN model, as well as the smaller standard deviation of FID. These results showcase the high potential of the new U-Net based discriminator.
For a detailed comparison of the FID mean, median and standard deviation across 5 runs we refer to Table \ref{table:fids_long} in the supplementary material. 

Qualitative results on FFHQ and COCO-Animals are shown in Figure \ref{fig:qual_results} and Figure \ref{fig:coco_pics}. Figure \ref{fig:qual_results} displays human faces generated by U-Net GAN through linear interpolation in the latent space between two synthetic samples. We observe that the interpolations are semantically smooth between faces, i.e. an open mouth gradually becomes a closed mouth, hair progressively grows in length, beards or glasses smoothly fade or appear, and hair color changes seamlessly. Furthermore, we notice that on several occasions men appear with pink beards. As FFHQ contains a fair share of people with pink hair, we suspect that our generator extrapolates hair color to beards, enabled by the global and local $D^U$ feedback during the training.
Figure \ref{fig:coco_pics} shows generated samples on COCO-Animals. We observe diverse images of high quality. We further notice that employing the class-conditional projection (as used in BigGAN) in the pixel output space of the decoder does not introduce class leakage or influence the class separation in any other way. These observations confirm that our U-Net GAN is effective in both unconditional and class-conditional image generation.

\begin{table}[t]
\vspace{-1em}
\setlength{\tabcolsep}{0.5em}
\renewcommand{\arraystretch}{1.1}
	\centering

	\begin{tabular}{l@{\hskip 1.3in}cc}
	Method & \normalsize{} FID $\downarrow$ & \normalsize{} IS $\uparrow$\tabularnewline
	\hline 	\hline

	\normalsize{} PG-GAN \cite{karras2018progressive} &  \normalsize{} 7.30 & \normalsize{} -- \tabularnewline
	\normalsize{} COCO-GAN~\cite{Lin2019COCOGANGB} &  \normalsize{} 5.74 & \normalsize{} -- \tabularnewline

	\normalsize{}  BigGAN~\cite{Brock2019} &  \normalsize{} 4.54 & \normalsize{} 3.23 \tabularnewline
	\normalsize{} U-Net GAN & \normalsize{} \textbf{2.95} & \normalsize{} \textbf{3.43} \tabularnewline
\end{tabular}

	\vspace{-0.5em}
    \caption{Comparison with the state-of-the-art models on CelebA ($128\times 128$). See Section~\ref{subsec:exp_results} for discussion.} \label{table:celeba_fid} %
	\vspace{-1em}
\end{table}

\paragraph{Ablation Study}

In Table \ref{table:fid_ablation} we next analyze the individual effect of each of the proposed components of the U-Net GAN model (see Section~\ref{method} for details) to the baseline architecture of BigGAN on the FFHQ and COCO-Animals datasets, comparing the median FID scores. Note that each of these individual components builds on each other. As shown in Table \ref{table:fid_ablation}, employing the U-Net architecture for the discriminator alone improves the median FID score from $12.42$ to $10.86$ for FFHQ and $16.55$ to $15.86$ for COCO-Animals. 
Adding the CutMix augmentation improves upon these scores even further, achieving FID of $10.30$ for FFHQ and $14.95$ for COCO-Animals. Note that we observe a similar improvement if we employ the CutMix augmentation during the BigGAN training as well.
Employing the proposed consistency regularization in the segmenter $D^U_{dec}$ output space on the CutMix images enables us to get the most out of the CutMix augmentation as well as allows to leverage better the per-pixel feedback of the U-Net discriminator, without imposing much computational or memory costs. In effect, the median FID score drops to $7.63$ for FFHQ and to $13.87$ for COCO-Animals.
Overall, we observe that each proposed component of the U-Net GAN model leads to improved performance in terms of FID.

\paragraph{Comparison with state of the art} 
Table \ref{table:celeba_fid} shows that U-Net GAN compares favourably with the state of the art on the CelebA dataset. 
The BigGAN baseline computed already outperforms COCO-GAN, the best result reported in the literature to the best of our knowledge, lowering FID from $5.74$ to $4.54$, whereas U-Net GAN further improves FID to $2.95$\footnote{FID scores for CelebA were computed with the standard TensorFlow Inception network for comparability. 
	The PyTorch and TensorFlow FIDs for all datasets are presented in Table \ref{table:supp_fid}.}. It is worth noting that BigGAN is the representative of just one of the two well known state-of-the art GAN families, led by BigGAN and StyleGAN, and their respective further improvements~\cite{Zhang2019ConsistencyRF,zhao2020improved,karras2019analyzing}. While in this paper we base our framework on BigGAN, it would be interesting to also explore the application of the U-Net based discriminator for the StyleGAN family. 

\begin{figure}[t]
	\begin{centering}
		\vspace{-2em}
		\hfill{}%
			\includegraphics[width=0.9\columnwidth]{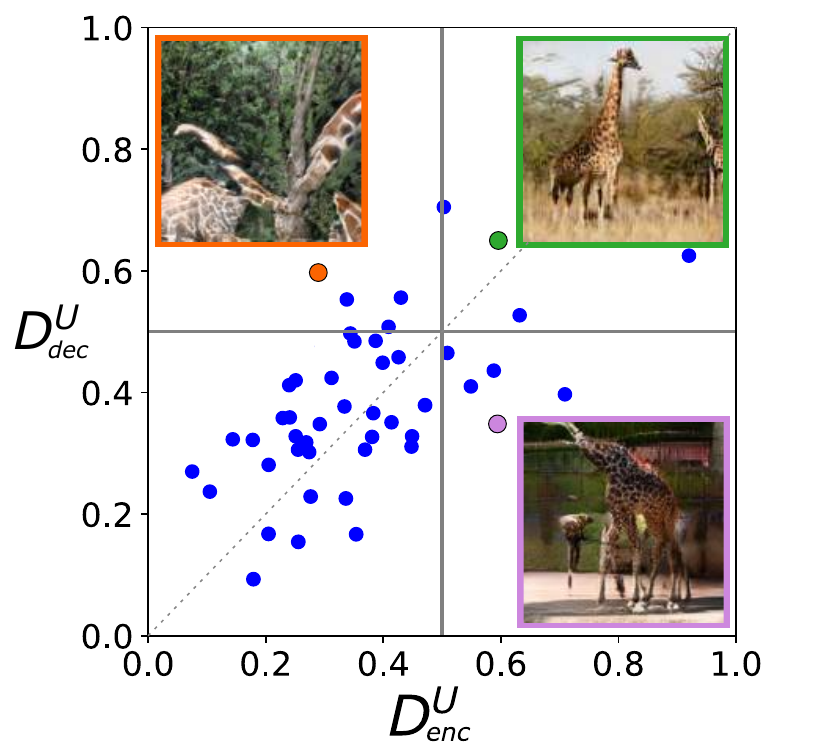}
     \hfill{}
		\par\end{centering}
	\vspace{-0.5em}
	\caption{\label{fig:disc_scores} Visualization of the predictions of the encoder $D^U_{enc}$ and decoder $D^U_{dec}$ modules during training, within a batch of $50$ generated samples. For visualization purposes, the $D^U_{dec}$ score is averaged over all pixels in the output. Note that quite often decisions of $D^U_{enc}$ and $D^U_{dec}$ are not coherent with each other. As judged by the U-Net discriminator, samples in the upper left consist of locally plausible patterns, while not being globally coherent (example in orange), whereas samples in the lower right look globally coherent but have local inconsistencies (example in purple: giraffe with too many legs and vague background). 
	}
	\vspace{-1em}
\end{figure}

\paragraph{Discriminator response visualization}
Experimentally we observe that $D^U_{enc}$ and $D^U_{dec}$ often assign different real/fake scores per sample. Figure \ref{fig:disc_scores} visualizes the per-sample predictions for a complete training batch. Here, the decoder score is computed as the average per-pixel prediction. The scores correlate with each other but have a high variance. Points in the upper left quadrant correspond to samples that are assigned a high probability of being real by the decoder, but a low probability by the encoder. This implies realism on a local level, but not necessarily on a global one. Similarly, the lower right quadrant represents samples that are identified as realistic by the encoder, but contain unrealistic patches which cause a low decoder score. The fact that the encoder and decoder predictions are not tightly coupled further implies that these two components are complementary. In other words, the generator receives more pronounced feedback by the proposed U-Net discriminator than it would get from a standard GAN discriminator. 

\paragraph{Characterizing the training dynamics}
\begin{figure}
\begin{centering}
\setlength{\tabcolsep}{0em}
\renewcommand{\arraystretch}{0}
\par\end{centering}
\begin{centering}
\vspace{-2em}
\hfill{}%
\includegraphics[width=1.0\columnwidth]{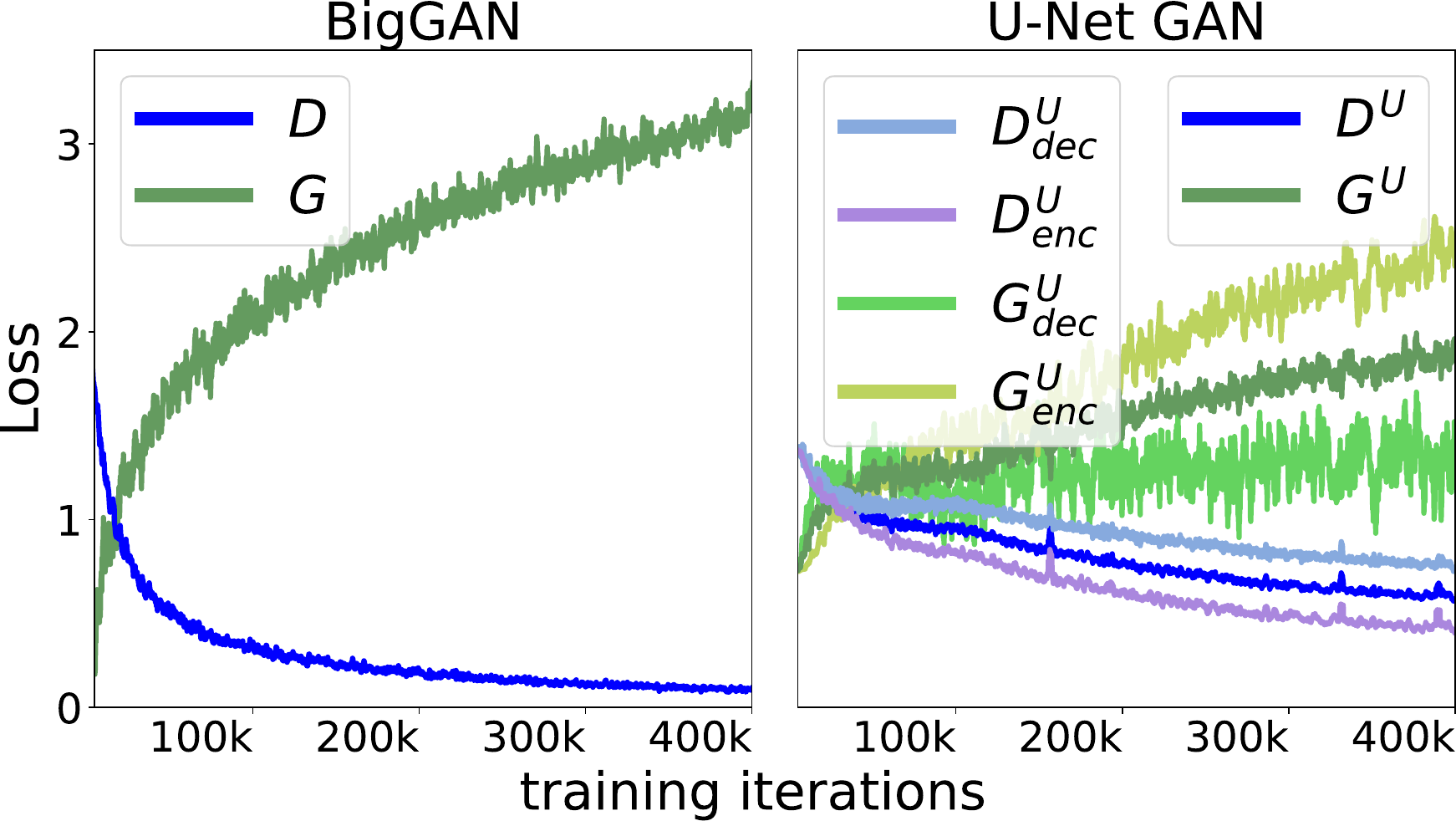}
\par\end{centering}
\vspace{-0.5em}
\caption{\label{fig:g_d_loss} Comparison of the generator and discriminator loss behavior over training for U-Net GAN and BigGAN. The generator and discriminator loss of U-Net GAN is additionally split up into its encoder- and decoder components.}	
\vspace{-1em}
\end{figure}

Both BigGAN and U-Net GAN experience similar stability issues, with $\sim60\%$ of all runs being successful.
For U-Net GAN, training collapse occurs generally much earlier ($\sim30$k iterations) than for BigGAN ($>200$k iterations, as also reported in~\cite{Brock2019}), allowing to discard failed runs earlier.
Among successful runs for both models, we observe a lower standard deviation in the achieved FID scores, compared to the BigGAN baseline (see Table \ref{table:fids_long} in the supplementary material). Figure \ref{fig:g_d_loss} depicts the evolution of the generator and discriminator losses (green and blue, respectively) for U-Net GAN and BigGAN over training. For U-Net GAN, the generator and discriminator losses are additionally split into the loss components of the U-Net encoder $D^U_{enc}$ and decoder $D^U_{dec}$. The U-Net GAN discriminator loss decays slowly, while the BigGAN discriminator loss approaches zero rather quickly, which prevents further learning from the generator. This explains the FID gains of U-Net GAN and shows its potential to improve with longer training. 
The generator and discriminator loss parts from encoder (image-level) and decoder (pixel-level) show similar trends, i.e. we observe the same decay for $D^U_{enc}$ and $D^U_{dec}$ losses but with different scales. 
This is expected as $D^U_{enc}$ can easily classify image as belonging to the real or fake class just by looking at one distinctive trait, while to achieve the same scale $D^U_{dec}$ needs to make a uniform real or fake decision on all image pixels.

\section{Conclusion}
In this paper, we propose an alternative U-Net based architecture for the discriminator, which allows to provide both global and local feedback to the generator. In addition, we introduce a consistency regularization technique for the U-Net discriminator based on the CutMix data augmentation. We show that all the proposed changes result in a stronger discriminator, enabling the generator to synthesize images with varying levels of detail, maintaining global and local realism. We demonstrate the improvement over the state-of-the-art BigGAN model~\cite{Brock2019} in terms of the FID score on three different datasets.

{\small
\bibliographystyle{ieee_fullname}
\bibliography{references}
}

\clearpage
\renewcommand{\thesubsection}{\Alph{subsection}}

\beginsupplement

\newpage
\section*{\Large Supplementary Material}
\label{content}

This supplementary material complements the presentation of U-Net GAN in the main paper with the following:

\begin{itemize}
\item Additional quantitative results in Section~\ref{sec:results};
\item Exemplar synthetic images on FFHQ in Section~\ref{sec:syn samples_ffhq} and on COCO-Animals in Section~\ref{sec:syn samples_coco}.
\item Network architectures and hyperparameter settings in Section~\ref{sec:networks}.	
\end{itemize}	

\subsection{Additional Evaluations}\label{sec:results}
Here we provide more detailed evaluation of the results presented in the main paper. In Table \ref{table:supp_fid} we report the inception metrics for images generated on FFHQ~\cite{Karras2018ASG}, COCO-Animals~\cite{Lin2014MicrosoftCC,OpenImages} and CelebA~\cite{Liu_Celeba} at resolution $256\times 256$, $128\times 128$, and $128\times 128$, respectively. In particular,
we report the Fr{\'e}chet Inception distance (FID)~\cite{heuselttur2017} and the Inception score (IS)~\cite{SalimansNeurIPS2016} computed by both the PyTorch\footnote{\url{https://github.com/ajbrock/BigGAN-PyTorch}} and TensorFlow\footnote{\url{https://github.com/bioinf-jku/TTUR}} implementations.
Note that the difference between two implementations lies in using either the TensorFlow or the PyTorch in-built inception network to calculate IS and FID, resulting in slightly different scores.
In all experiments, FID and IS are computed using $50$\unit{k} synthetic images, following~\cite{karras2018progressive}.
By default all reported numbers correspond to the best FID achieved with $400$\unit{k} training iterations for FFHQ and COCO-Animals, and $800$\unit{k} iterations for CelebA, using the PyTorch implementation. 

In the unconditional case, on FFHQ, our model achieves FID of $7.48$ ($8.88$ in TensorFlow), which is an improvement of $4.0$ ($6.04$ in TensorFlow) FID points over the BigGAN discriminator \cite{Brock2019}. 
The same effect is observed for the conditional image generation setting on COCO-Animals. Here, our U-Net GAN achieves FID of $13.73$ ($13.96$ in TensorFlow), improving $2.64$ ($2.46$ in TensorFlow) points over BigGAN. 
To compare with other state-of-the-art models we additionally evaluate U-Net GAN on CelebA for unconditional image synthesis. Our U-Net GAN achieves $2.95$ FID (in TensorFlow), outperforming COCO-GAN~\cite{Lin2019COCOGANGB}, PG-GAN \cite{karras2018progressive}, and the BigGAN baseline \cite{Brock2019}. 

Table \ref{table:fids_long} shows that U-Net GAN does not only outperform the BigGAN baseline in terms of the best recorded FID, but also with respect to the mean, median and standard deviation computed over 5 independent runs. Note the strong drop in standard deviation from $0.24$ to $0.11$ on COCO-Animals and from $0.16$ to $0.04$ on CelebA.

\begin{table}[t]
	\vspace{0em}
	\setlength{\tabcolsep}{0.25em}
	\renewcommand{\arraystretch}{1.5}
	\centering

	\resizebox{\columnwidth}{!}{
	\begin{tabular}{c@{\hskip 0.1in}lcc@{\hskip 0.2in}cc}

		&  & \multicolumn{2}{c@{\hskip 0.2in}}{\normalsize{} PyTorch}	& \multicolumn{2}{c}{\normalsize{}{TensorFlow}} \tabularnewline
		Dataset & Method  & \normalsize{} FID $\downarrow$ & \normalsize{} IS $\uparrow$  & \normalsize{} FID $\downarrow$ & \normalsize{} IS $\uparrow$\tabularnewline
		\hline 	\hline
		FFHQ & \normalsize{} BigGAN~\cite{Brock2019}  & \normalsize{} 11.48	& \normalsize{} 3.97 & \normalsize{} 14.92 & \normalsize{} 3.96  \tabularnewline
		  ($256\times 256$)  & \normalsize{} U-Net GAN & \normalsize{} \textbf{7.48}	& \normalsize{}  \textbf{4.46} &  \normalsize{} \textbf{8.88} & \normalsize{} \textbf{4.50} \tabularnewline
		  \hline
		COCO-Animals & \normalsize{} BigGAN~\cite{Brock2019}  & \normalsize{} 16.37	& \normalsize{}  11.77 &  \normalsize{} 16.42 & \normalsize{} 11.34 \tabularnewline
		   	($128\times 128$)		& \normalsize{} U-Net GAN & \normalsize{} \textbf{13.73}& \normalsize{}  \textbf{12.29} &  \normalsize{} \textbf{13.96} & \normalsize{} \textbf{11.77} \tabularnewline
		   			  \hline
		& \normalsize{} PG-GAN \cite{karras2018progressive}  & \normalsize{} --	& \normalsize{}  -- &  \normalsize{} 7.30 & \normalsize{} -- \tabularnewline
		CelebA & \normalsize{} COCO-GAN~\cite{Lin2019COCOGANGB}  & \normalsize{} --	& \normalsize{}  -- &  \normalsize{} 5.74 & \normalsize{} -- \tabularnewline
		($128\times 128$) & \normalsize{}  BigGAN~\cite{Brock2019}  & \normalsize{} 3.70	& \normalsize{}  3.08 &  \normalsize{} 4.54 & \normalsize{} 3.23 \tabularnewline
		  	              & \normalsize{} U-Net GAN & \normalsize{} \textbf{2.03}	& \normalsize{}  \textbf{3.33} &  \normalsize{} \textbf{2.95} & \normalsize{} \textbf{3.43} \tabularnewline
	\end{tabular}
	}

	\vspace{0em}
    \caption{Evaluation results on FFHQ, COCO-Animals and CelebA with PyTorch and TensorFlow FID/IS scores. The difference lies in the choice of framework in which the inception network is implemented, which is used to extract the inception metrics. See Section~\ref{sec:results} for discussion.} \label{table:supp_fid} %
	\vspace{0em}
\end{table}

\begin{table}[t]
	\vspace{-0em}
	\setlength{\tabcolsep}{0.5em}
	\renewcommand{\arraystretch}{0.8}
	\centering

\begin{tabular}{l c c c  c  c }
    \multirow{2}{*}{\footnotesize Method}    & \multirow{2}{*}{ \footnotesize Dataset} & \multicolumn{4}{c}{ \footnotesize FID} \\
       &  &  \footnotesize Best &  \footnotesize Median &  \footnotesize Mean  &  \footnotesize Std \\
    	\hline

  	\footnotesize BigGAN    & \multirow{2}{*}{ \footnotesize COCO-Animals} & \footnotesize 16.37 & \footnotesize 16.55 & \footnotesize 16.62 & \footnotesize 0.24 \\
     	\footnotesize U-Net GAN &  & \footnotesize  \textbf{13.73} & \footnotesize \textbf{13.87} & \footnotesize \textbf{13.88} & \footnotesize \textbf{0.11} \\
    	\hline
    	\footnotesize BigGAN    &  \multirow{2}{*}{ \footnotesize FFHQ}   & \footnotesize  11.48 & \footnotesize 12.42 & \footnotesize 12.35 & \footnotesize 0.67 \\
     	\footnotesize U-Net GAN &   & \footnotesize  \textbf{7.48}  & \footnotesize \textbf{7.63}  & \footnotesize \textbf{7.73}  & \footnotesize \textbf{0.56} \\
	\hline
      	\footnotesize BigGAN    &  \multirow{2}{*}{ \footnotesize CelebA}   & \footnotesize  3.70 & \footnotesize 3.89 & \footnotesize 3.94 & \footnotesize 0.16 \\
      	\footnotesize U-Net GAN &   & \footnotesize  \textbf{2.03}  & \footnotesize \textbf{2.07}  & \footnotesize \textbf{2.08}  & \footnotesize \textbf{0.04} \\
	\end{tabular}
	\caption{ Best, median, mean and std of FID values across 5 runs.} \label{table:fids_long}
	\vspace{-1.5em}
\end{table}

\subsection{Qualitative Results on FFHQ}\label{sec:syn samples_ffhq}
Here we present more qualitative results of U-Net GAN on FFHQ \cite{Karras2018ASG}.
We use FFHQ for unconditional image synthesis and generate images with a resolution of $256\times 256$.

\subsubsection*{Generated FFHQ samples}
\begin{figure*}
	\begin{centering}
\vspace{-1em}
    \includegraphics[width=0.9\textwidth]{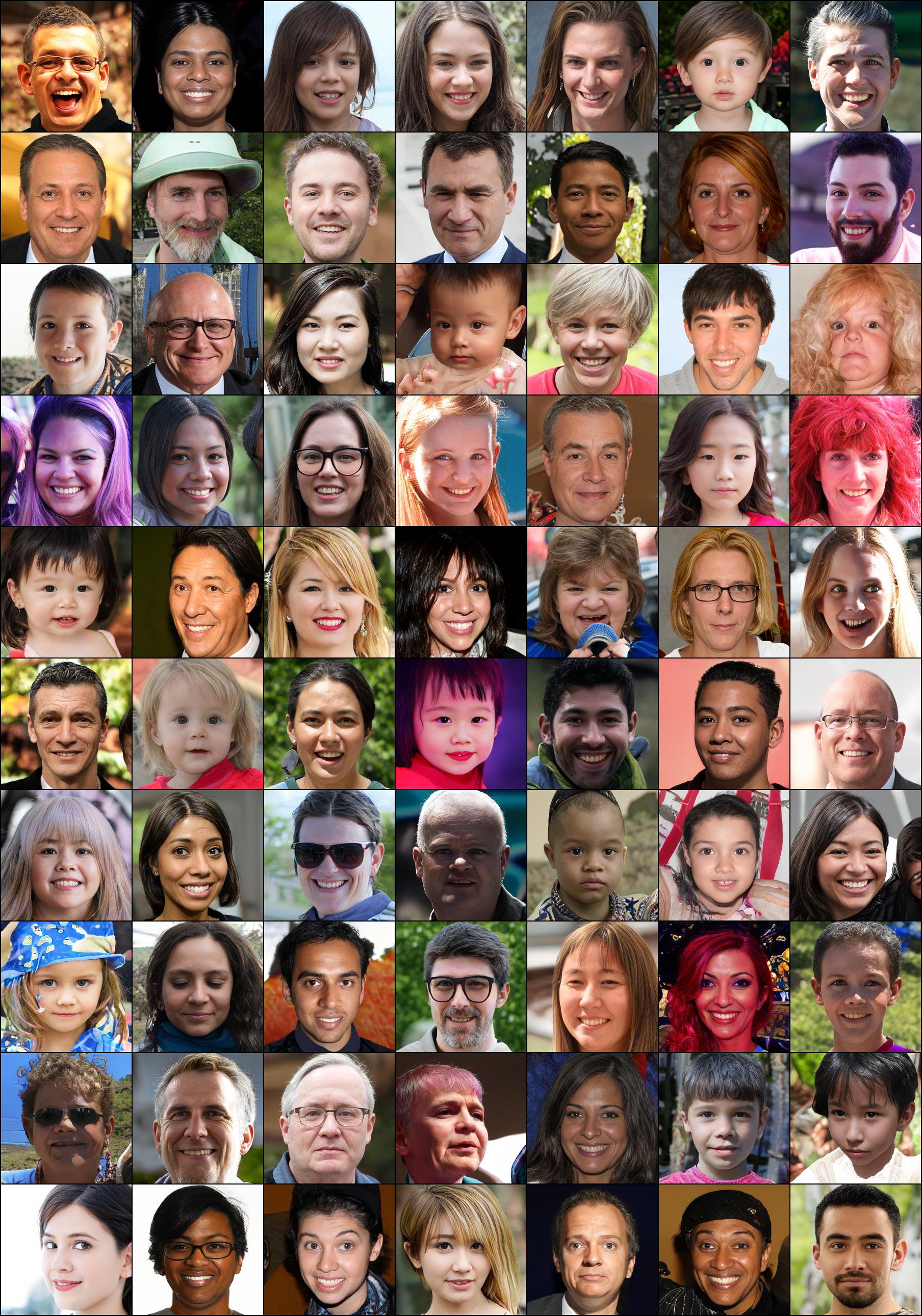}    	
	\caption{\label{fig:random_ffhq_samples} Images generated by U-Net GAN trained on FFHQ with resolution $256 \times 256$.}
\end{centering}
\end{figure*}

Figure \ref{fig:random_ffhq_samples} shows samples of human faces generated by U-Net GAN on FFHQ. We observe diverse images of high quality, maintaining local and global realism.

\subsubsection*{Per-pixel U-Net discriminator feedback}
\begin{figure*}
	\begin{centering}
		\setlength{\tabcolsep}{0.1em}
		\renewcommand{\arraystretch}{0}
		\par\end{centering}
	\begin{centering}
		\begin{tabular}{@{}c@{\hskip 0.05in }c@{\hskip 0.05in }c@{\hskip 0.05in }c@{\hskip 0.05in }c@{}}

			{\footnotesize{}}\includegraphics[width=0.19\textwidth]{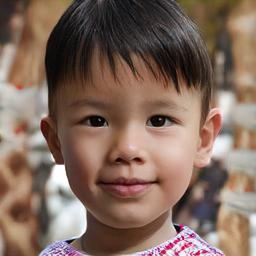} &
			{\footnotesize{}}\includegraphics[width=0.19\textwidth]{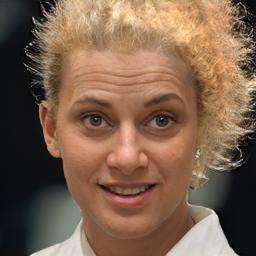} &
			{\footnotesize{}}\includegraphics[width=0.19\textwidth]{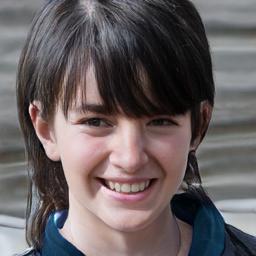} &
			{\footnotesize{}}\includegraphics[width=0.19\textwidth]{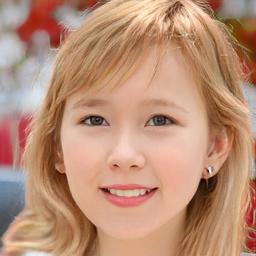} &
			{\footnotesize{}}\includegraphics[width=0.19\textwidth]{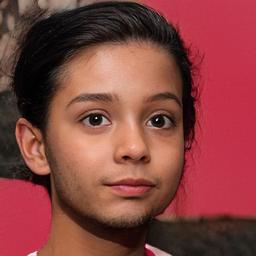} \\

			{\footnotesize{}}\includegraphics[width=0.19\textwidth]{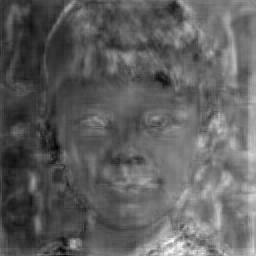} &
			{\footnotesize{}}\includegraphics[width=0.19\textwidth]{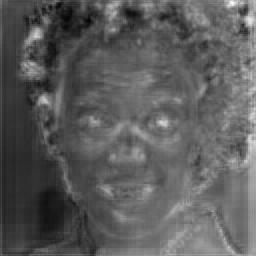} &
			{\footnotesize{}}\includegraphics[width=0.19\textwidth]{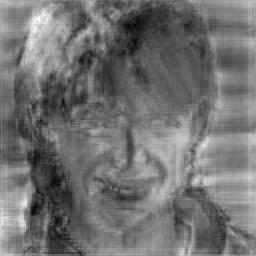} &
			{\footnotesize{}}\includegraphics[width=0.19\textwidth]{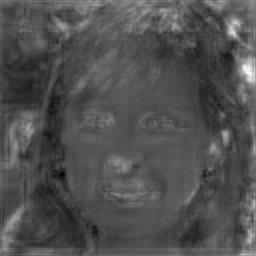} &
			{\footnotesize{}}\includegraphics[width=0.19\textwidth]{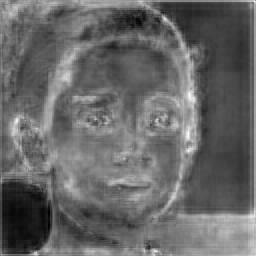} \\

			{\footnotesize{}}\includegraphics[width=0.19\textwidth]{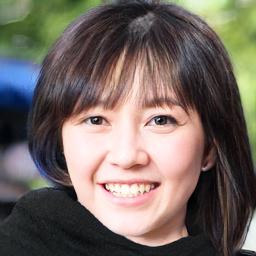} &
			{\footnotesize{}}\includegraphics[width=0.19\textwidth]{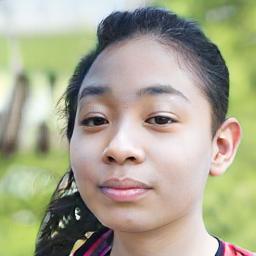} &
			{\footnotesize{}}\includegraphics[width=0.19\textwidth]{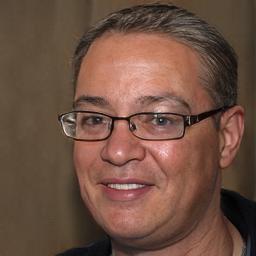} &
			{\footnotesize{}}\includegraphics[width=0.19\textwidth]{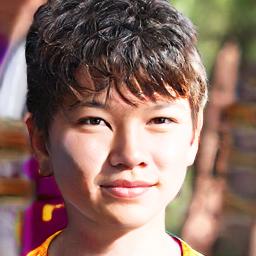} &
			{\footnotesize{}}\includegraphics[width=0.19\textwidth]{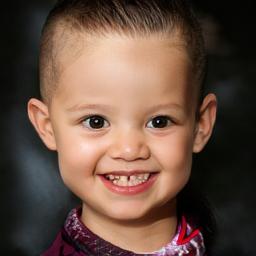} \\

			{\footnotesize{}}\includegraphics[width=0.19\textwidth]{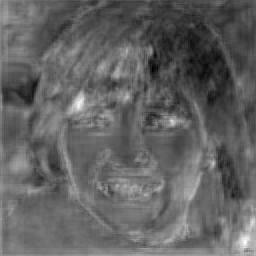} &
			{\footnotesize{}}\includegraphics[width=0.19\textwidth]{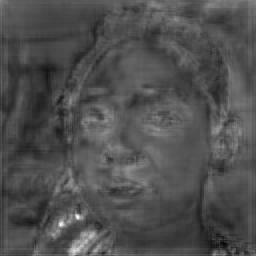} &
			{\footnotesize{}}\includegraphics[width=0.19\textwidth]{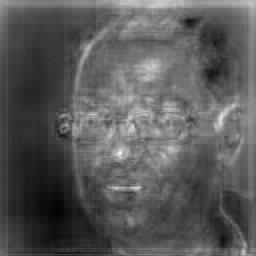} &
			{\footnotesize{}}\includegraphics[width=0.19\textwidth]{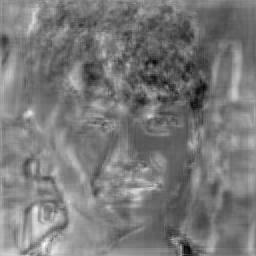} &
			{\footnotesize{}}\includegraphics[width=0.19\textwidth]{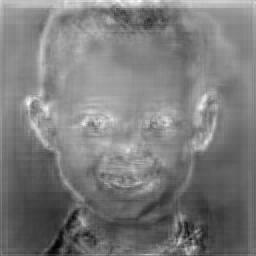}\\

			{\footnotesize{}}\includegraphics[width=0.19\textwidth]{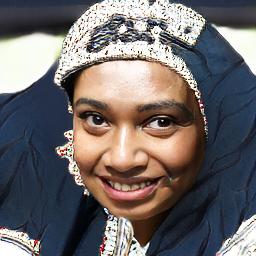} &
			{\footnotesize{}}\includegraphics[width=0.19\textwidth]{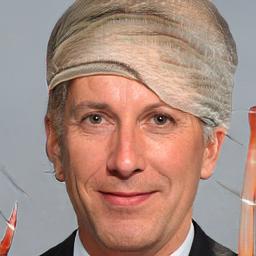} &
				{\footnotesize{}}\includegraphics[width=0.19\textwidth]{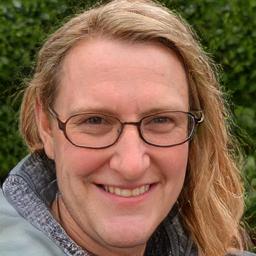} &
				{\footnotesize{}}\includegraphics[width=0.19\textwidth]{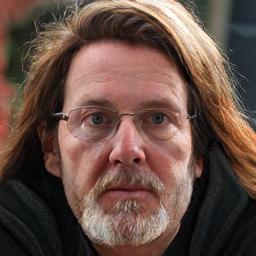} &
			{\footnotesize{}}\includegraphics[width=0.19\textwidth]{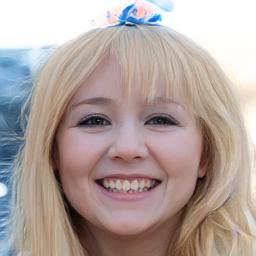} \\

			{\footnotesize{}}\includegraphics[width=0.19\textwidth]{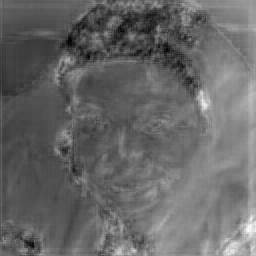} &
			{\footnotesize{}}\includegraphics[width=0.19\textwidth]{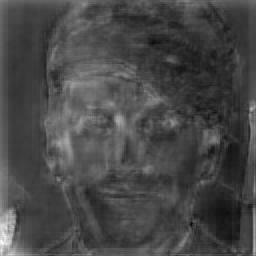} &
				{\footnotesize{}}\includegraphics[width=0.19\textwidth]{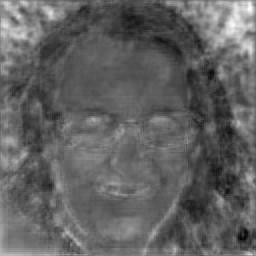} &
			{\footnotesize{}}\includegraphics[width=0.19\textwidth]{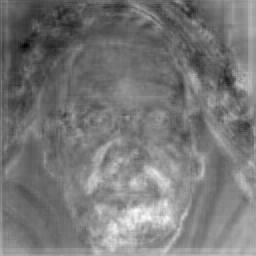}&
			{\footnotesize{}}\includegraphics[width=0.19\textwidth]{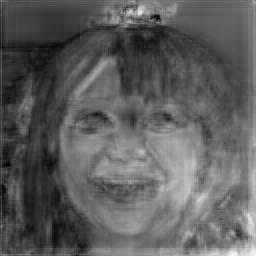} \\

		\end{tabular}

		\par\end{centering}
	\caption{\label{fig:maps} Samples generated by U-Net GAN and the corresponding real-fake predictions of the U-Net decoder. Brighter colors correspond to the discriminator confidence of pixel being real (and darker of being fake).}
\end{figure*}

In Figure \ref{fig:maps} we visualize synthetic images and their corresponding per-pixel feedback of the U-Net discriminator. Note that the U-Net discriminator provides a very detailed and spatially coherent response, which enables the generator to further improve the image quality.

\subsubsection*{Interpolations in the latent space}
\begin{figure*}
\begin{centering}
\setlength{\tabcolsep}{0.1em}
\renewcommand{\arraystretch}{0}
\par\end{centering}
\begin{centering}
\vspace{-1.5em}
\hfill{}%
\begin{tabular}{c@{\hskip 0.05in}c@{\hskip 0.05in}c@{\hskip 0.05in}c@{\hskip 0.05in}c@{\hskip 0.05in}c}

\includegraphics[width=0.85\textwidth]{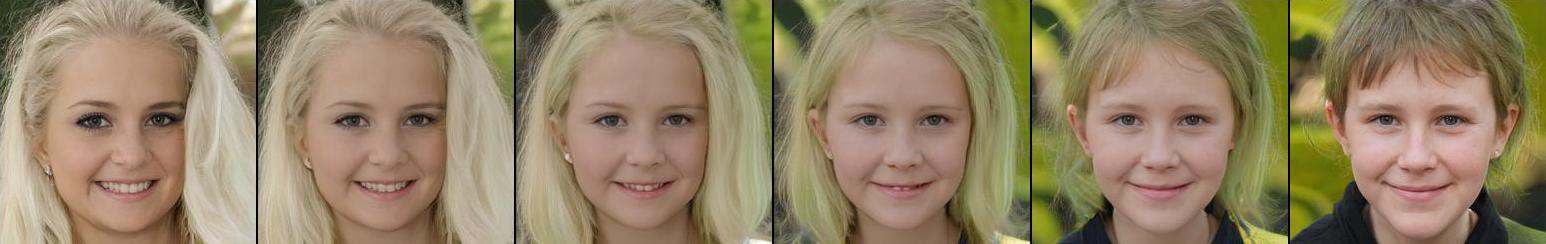}\tabularnewline
\includegraphics[width=0.85\textwidth]{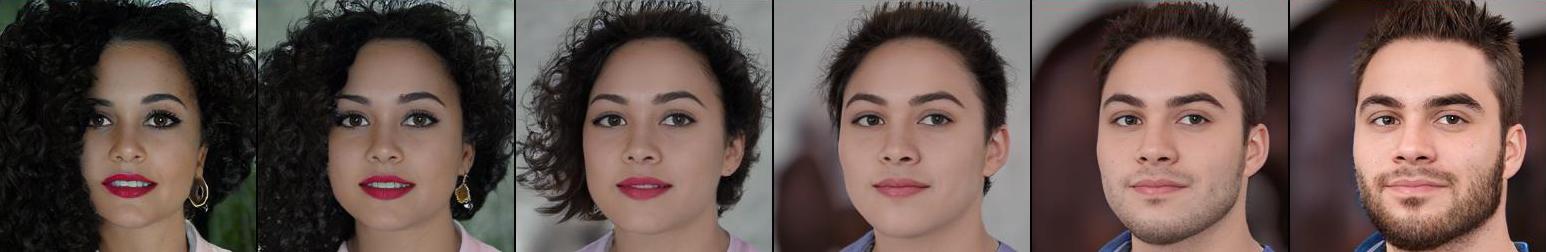}\tabularnewline
\includegraphics[width=0.85\textwidth]{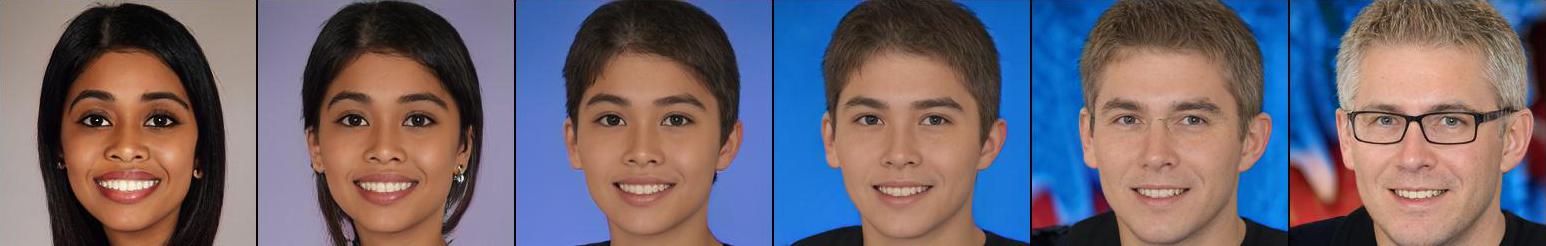}\tabularnewline
\includegraphics[width=0.85\textwidth]{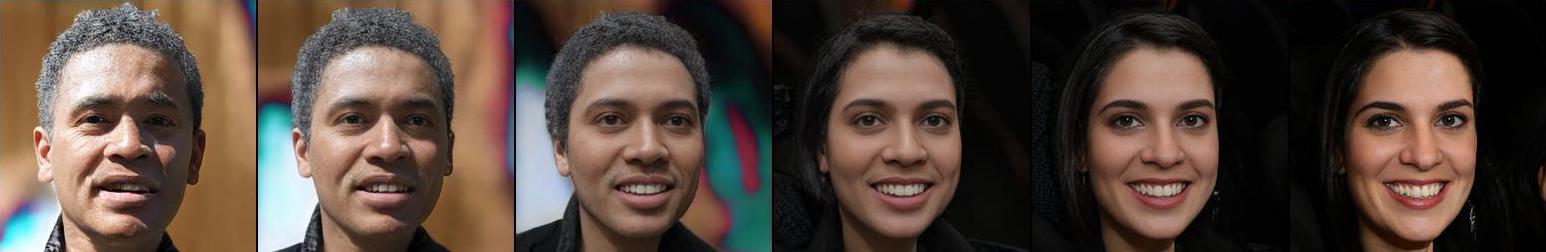}\tabularnewline
\includegraphics[width=0.85\textwidth]{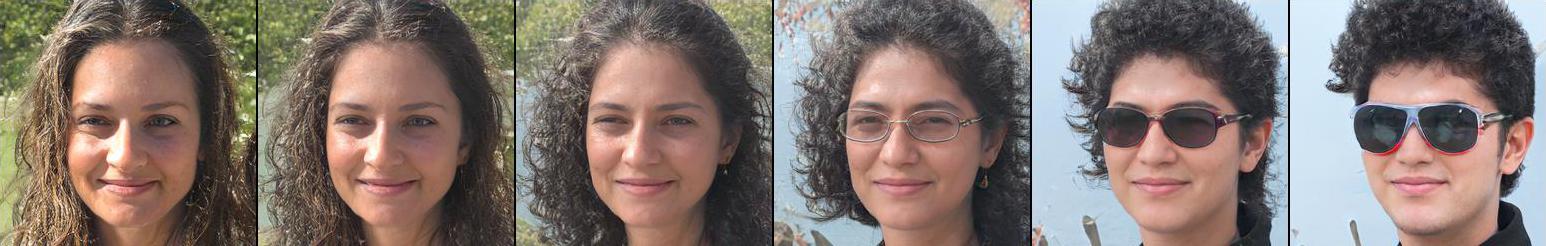}\tabularnewline

\includegraphics[width=0.85\textwidth]{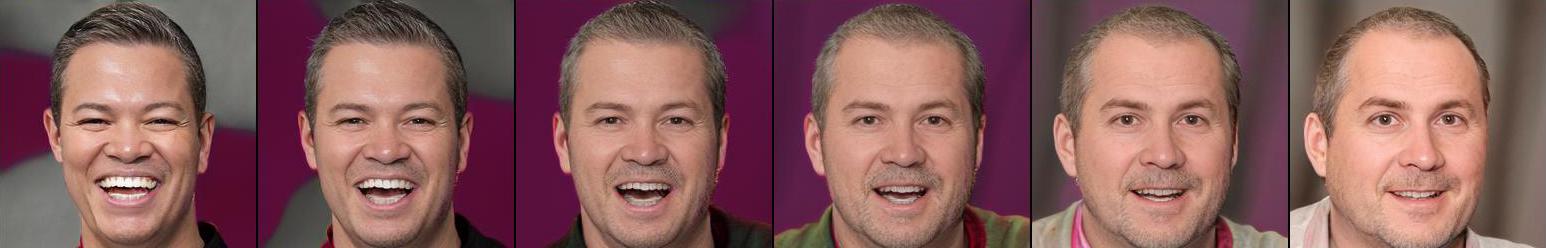}\tabularnewline
\includegraphics[width=0.85\textwidth]{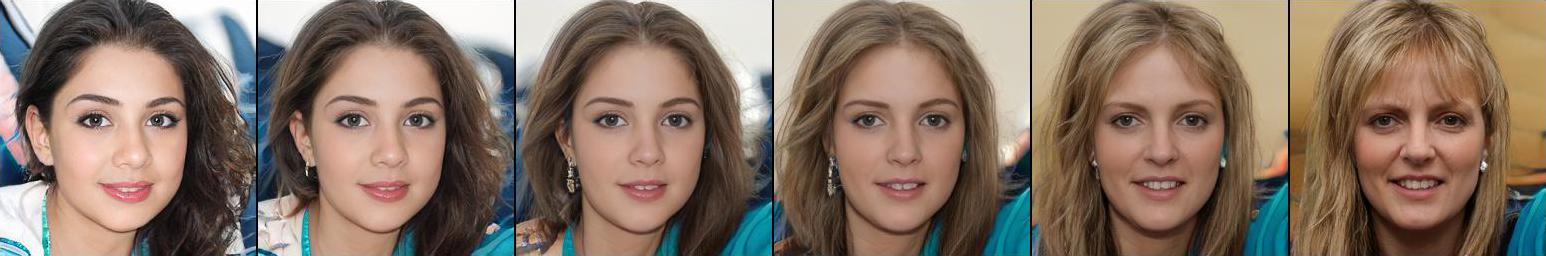}\tabularnewline
\includegraphics[width=0.85\textwidth]{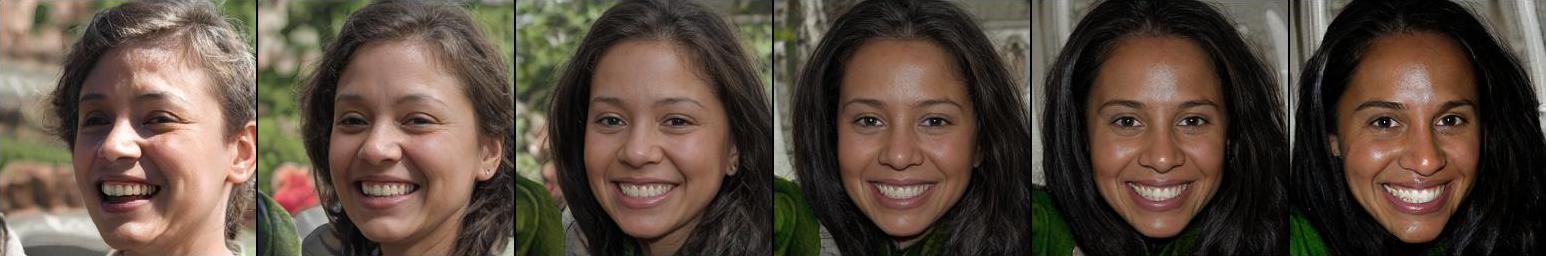}\tabularnewline
\includegraphics[width=0.85\textwidth]{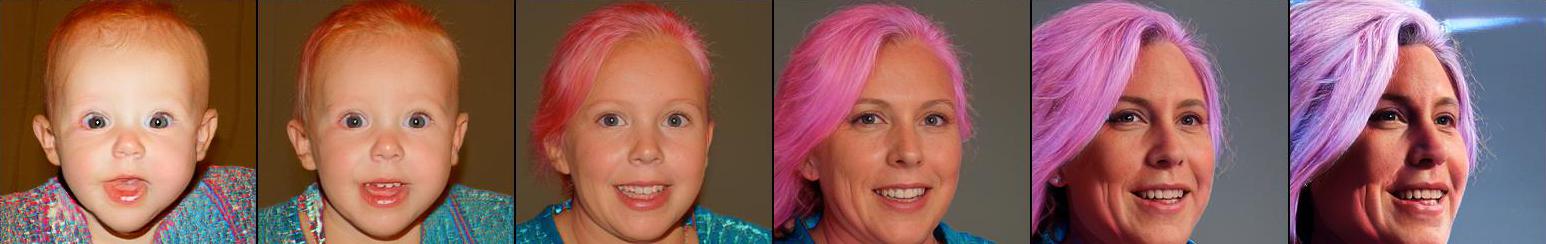}\tabularnewline

\end{tabular}\hfill{}
\par\end{centering}
\vspace{-0.5em}
\caption{\label{fig:more_ffhq_interpolations} Images generated with U-Net GAN on FFHQ with resolution $256 \times 256$ when interpolating in the latent space.} %
\vspace{0em}
\end{figure*}

Figure \ref{fig:more_ffhq_interpolations} displays human faces generated by U-Net GAN through linear interpolation in the latent space between two synthetic samples. We observe that the interpolations are semantically smooth between faces, e.g. an open mouth gradually becomes a closed mouth, hair progressively grows or gets shorter in length, beards or glasses smoothly fade or appear, and hair color changes seamlessly. 

\subsubsection*{Comparison between BigGAN and U-Net GAN}
\begin{figure*}
\begin{centering}
\setlength{\tabcolsep}{0.1em}
\renewcommand{\arraystretch}{1.0}
\par\end{centering}
\begin{centering}
\hfill{}%
\begin{tabular}{c@{\hskip 0.05in}c@{\hskip 0.05in}c@{\hskip 0.05in}c@{\hskip 0.05in}c@{\hskip 0.05in}c}
 
BigGAN \tabularnewline
\includegraphics[width=0.85\textwidth]{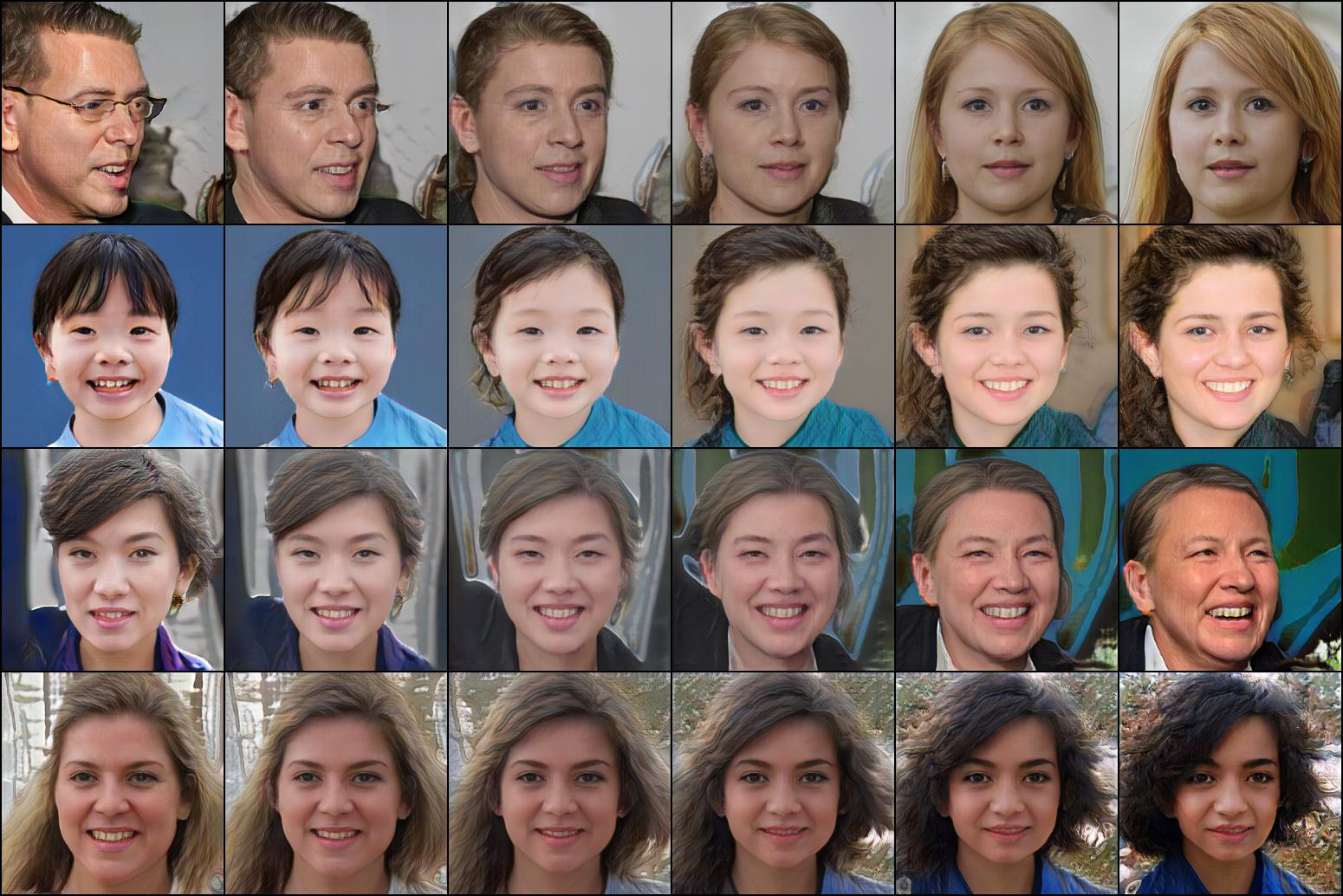}\tabularnewline
\tabularnewline
U-Net GAN \tabularnewline
\includegraphics[width=0.85\textwidth]{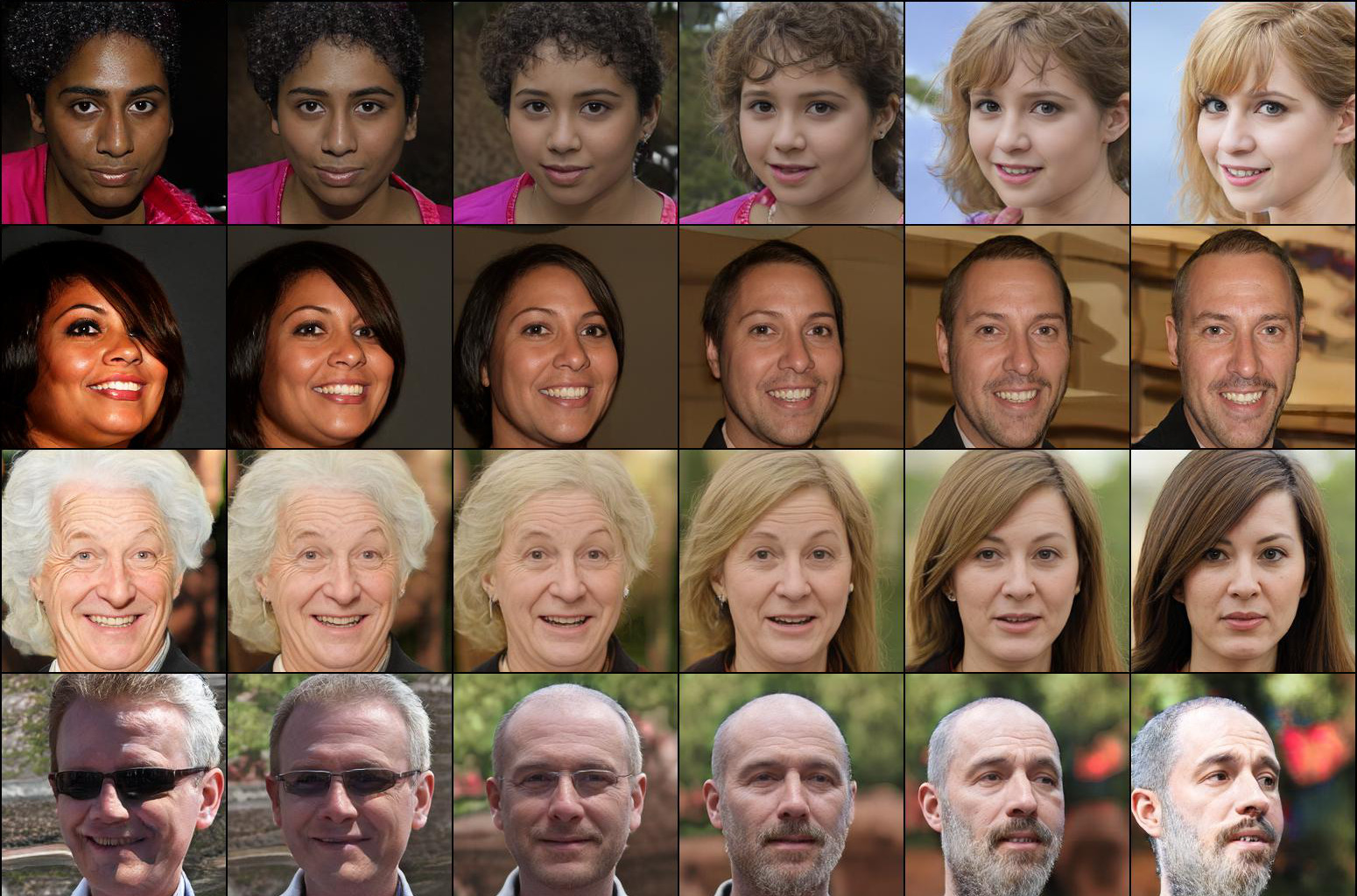}

\end{tabular}\hfill{}
\par\end{centering}
\caption{\label{fig:qual_comparison} Qualitative comparison of uncurated images generated with the unconditional BigGAN model (top) and our U-Net GAN (bottom) on FFHQ with resolution $256 \times 256$. Note that the images generated by U-Net GAN exhibit finer details and maintain better local realism.} %
\vspace{0em}
\end{figure*}

In Figure \ref{fig:qual_comparison} we present a qualitative comparison of uncurated images generated with the unconditional BigGAN model~\cite{Brock2019} and our U-Net GAN. Note that the images generated by U-Net GAN exhibit finer details and maintain better local realism.

\subsubsection*{CutMix images and U-Net discriminator predictions}
\begin{figure*}
		\setlength{\tabcolsep}{0em}
		\renewcommand{\arraystretch}{1.0}
		\vspace{-1em}

	\begin{centering}

\begin{tabular}{@{\hskip -0.2in}c@{ }c@{ }c@{ }c@{ }c@{ }c@{}c@{ }c@{ }c@{ }c@{ }c@{}}

	 & \multicolumn{2}{c}{\footnotesize{} Real}  & \multicolumn{2}{c}{\footnotesize{} Fake} &  & \multicolumn{2}{c}{\footnotesize{} Real}  & \multicolumn{2}{c}{\footnotesize{} Fake}   \\
	  \vspace{0.5em}
  \begin{minipage}{0.19\columnwidth} 	\hfill{} \begin{tabular}{c} {	\footnotesize{} Original} \\
{\footnotesize{} images}\end{tabular} 	\hfill{} \end{minipage}  &
 \begin{minipage}{0.2\columnwidth} 	\includegraphics[width=1\columnwidth, height=1\columnwidth]{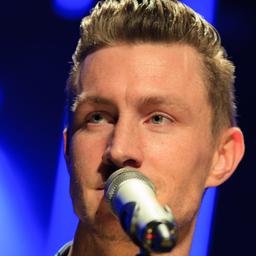}  \end{minipage} &
  \begin{minipage}{0.2\columnwidth} 	\includegraphics[width=1\columnwidth, height=1\columnwidth]{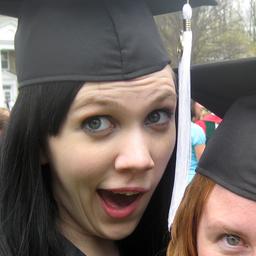}  \end{minipage} &
 \begin{minipage}{0.2\columnwidth} 	\includegraphics[width=1\columnwidth, height=1\columnwidth]{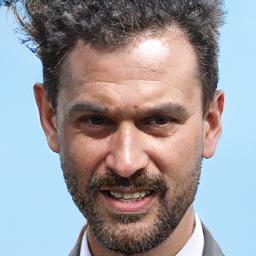}  \end{minipage} &
 \begin{minipage}{0.2\columnwidth} 	\includegraphics[width=1\columnwidth, height=1\columnwidth]{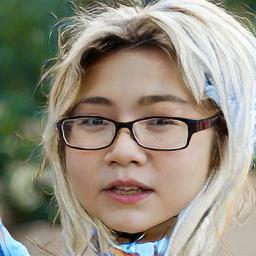}  \end{minipage} &
 \begin{minipage}{0.1\columnwidth} $\ \ \ \ \ \ \ $ \end{minipage} &
\begin{minipage}{0.2\columnwidth} 	\includegraphics[width=1\columnwidth, height=1\columnwidth]{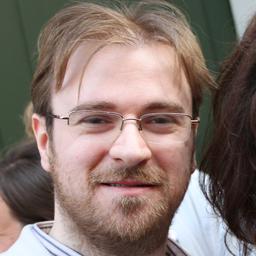}  \end{minipage} &
\begin{minipage}{0.2\columnwidth} 	\includegraphics[width=1\columnwidth, height=1\columnwidth]{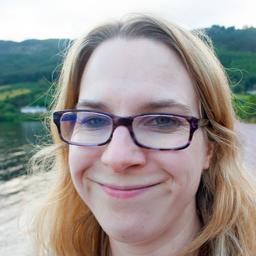}  \end{minipage} &
\begin{minipage}{0.2\columnwidth} 	\includegraphics[width=1\columnwidth, height=1\columnwidth]{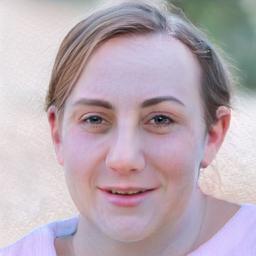}  \end{minipage} &
\begin{minipage}{0.2\columnwidth} 	\includegraphics[width=1\columnwidth, height=1\columnwidth]{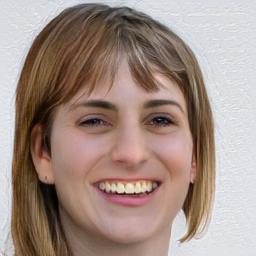}  \end{minipage}\\

\vspace{0.0em}
\begin{minipage}{0.19\columnwidth} 	\hfill{} \begin{tabular}{c}	\footnotesize{}Real/fake \\
		\footnotesize{} ratio $r$ 	\end{tabular} 	\hfill{} \end{minipage} & 0.86 & 0.58 &  0.51 &  0.70 & & 0.61 & 0.84 &  0.97 &  0.85 \\
\vspace{0.5em}

\begin{minipage}{0.19\columnwidth}	\hfill{}	\begin{tabular}{c}	\footnotesize{} Mask \\
		\footnotesize{} $\mathrm{M}$ \end{tabular} 	\hfill{} \end{minipage} &
\begin{minipage}{0.2\columnwidth}	\includegraphics[width=1\columnwidth, height=1\columnwidth, cfbox=verylightgray 0.2pt 0.2pt]{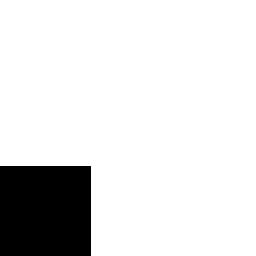} \end{minipage}&
\begin{minipage}{0.2\columnwidth}	\includegraphics[width=1\columnwidth, height=1\columnwidth, cfbox=verylightgray 0.2pt 0.2pt]{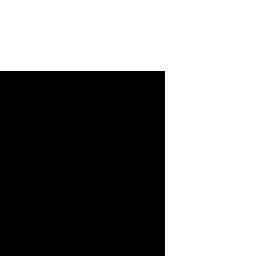} \end{minipage} &
\begin{minipage}{0.2\columnwidth}	\includegraphics[width=1\columnwidth, height=1\columnwidth, cfbox=verylightgray 0.2pt 0.2pt]{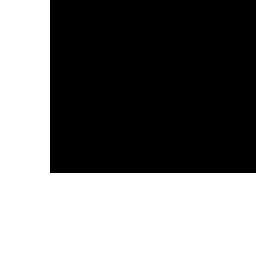} \end{minipage} &
\begin{minipage}{0.2\columnwidth}	\includegraphics[width=1\columnwidth, height=1\columnwidth, cfbox=verylightgray 0.2pt 0.2pt]{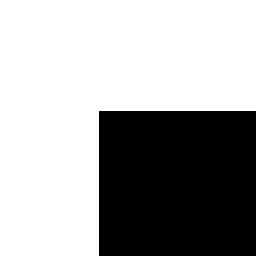} \end{minipage} &
 \begin{minipage}{0.1\columnwidth} $\ \ \ \ \ \ \ $ \end{minipage} &
\begin{minipage}{0.2\columnwidth}	\includegraphics[width=1\columnwidth, height=1\columnwidth, cfbox=verylightgray 0.2pt 0.2pt]{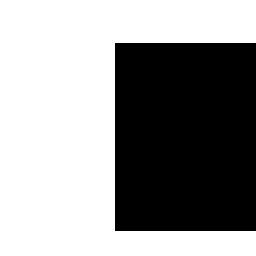} \end{minipage}&
\begin{minipage}{0.2\columnwidth}	\includegraphics[width=1\columnwidth, height=1\columnwidth, cfbox=verylightgray 0.2pt 0.2pt]{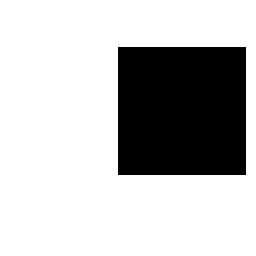} \end{minipage} &
\begin{minipage}{0.2\columnwidth}	\includegraphics[width=1\columnwidth, height=1\columnwidth, cfbox=verylightgray 0.2pt 0.2pt]{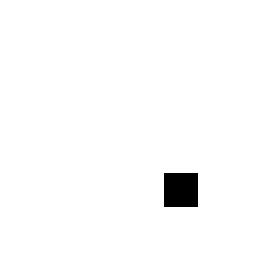} \end{minipage} &
\begin{minipage}{0.2\columnwidth}	\includegraphics[width=1\columnwidth, height=1\columnwidth, cfbox=verylightgray 0.2pt 0.2pt]{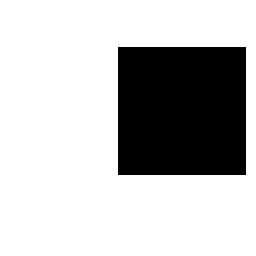} \end{minipage}\\
\vspace{0.5em}

  \begin{minipage}{0.19\columnwidth}	\hfill{}	\begin{tabular}{c}	\footnotesize{} CutMix \\
	\footnotesize{} images \end{tabular} 	\hfill{} \end{minipage} &
  \begin{minipage}{0.2\columnwidth}	\includegraphics[width=1\columnwidth, height=1\columnwidth]{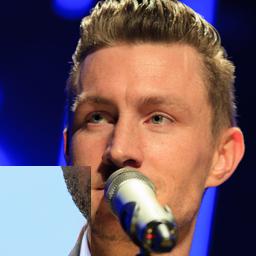} \end{minipage}&
   \begin{minipage}{0.2\columnwidth}	\includegraphics[width=1\columnwidth, height=1\columnwidth]{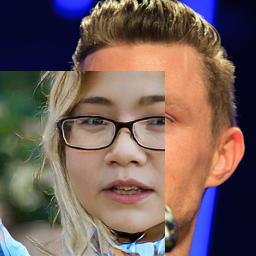} \end{minipage} &
  \begin{minipage}{0.2\columnwidth}	\includegraphics[width=1\columnwidth, height=1\columnwidth]{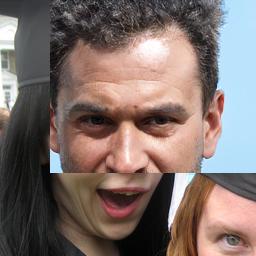} \end{minipage} &
  \begin{minipage}{0.2\columnwidth}	\includegraphics[width=1\columnwidth, height=1\columnwidth]{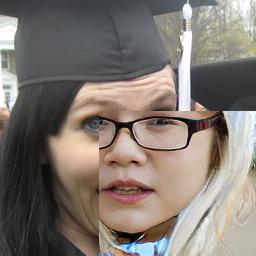} \end{minipage} &
 \begin{minipage}{0.1\columnwidth} $\ \ \ \ \ \ \ $ \end{minipage} &
\begin{minipage}{0.2\columnwidth}	\includegraphics[width=1\columnwidth, height=1\columnwidth]{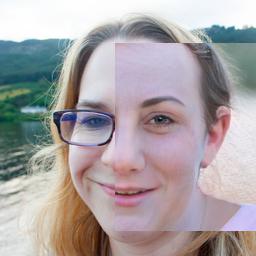} \end{minipage}&
\begin{minipage}{0.2\columnwidth}	\includegraphics[width=1\columnwidth, height=1\columnwidth]{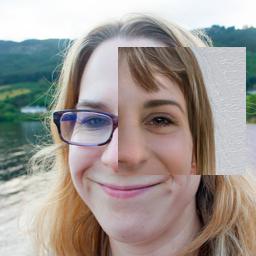} \end{minipage}&
\begin{minipage}{0.2\columnwidth}	\includegraphics[width=1\columnwidth, height=1\columnwidth]{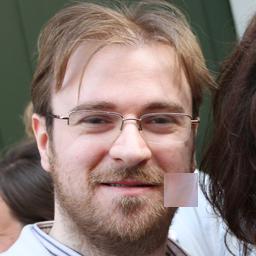} \end{minipage}&
\begin{minipage}{0.2\columnwidth}	\includegraphics[width=1\columnwidth, height=1\columnwidth]{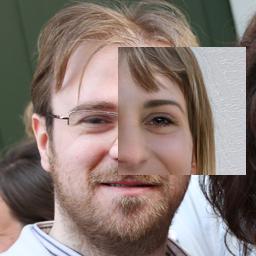} \end{minipage}\\
\vspace{0.5em}

   \begin{minipage}{0.19\columnwidth} 	\hfill{} \begin{tabular}{c}	\footnotesize{} $D^U_{dec}$ segm. \\
	\footnotesize{} map 	\end{tabular} 	\hfill{} \end{minipage} &
\begin{minipage}{0.2\columnwidth} \includegraphics[width=1\columnwidth, height=1\columnwidth]{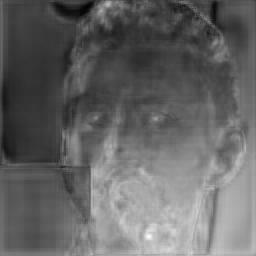} \end{minipage} &
\begin{minipage}{0.2\columnwidth} \includegraphics[width=1\columnwidth, height=1\columnwidth]{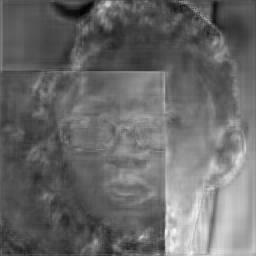} \end{minipage} &
\begin{minipage}{0.2\columnwidth} \includegraphics[width=1\columnwidth, height=1\columnwidth]{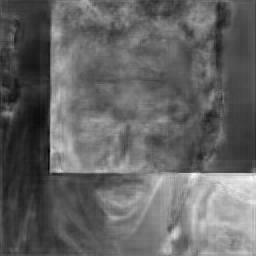}  \end{minipage} &
  \begin{minipage}{0.2\columnwidth} \includegraphics[width=1\columnwidth, height=1\columnwidth]{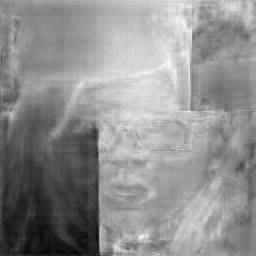} \end{minipage} &
 \begin{minipage}{0.1\columnwidth} $\ \ \ \ \ \ \ $ \end{minipage} &
\begin{minipage}{0.2\columnwidth} \includegraphics[width=1\columnwidth, height=1\columnwidth]{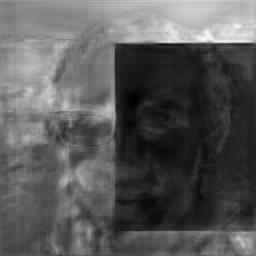} \end{minipage} &
\begin{minipage}{0.2\columnwidth} \includegraphics[width=1\columnwidth, height=1\columnwidth]{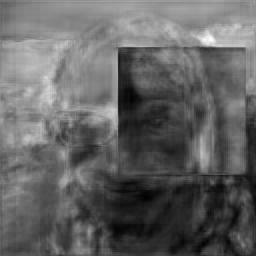} \end{minipage} &
\begin{minipage}{0.2\columnwidth} \includegraphics[width=1\columnwidth, height=1\columnwidth]{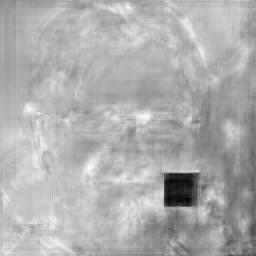} \end{minipage} &
\begin{minipage}{0.2\columnwidth} \includegraphics[width=1\columnwidth, height=1\columnwidth]{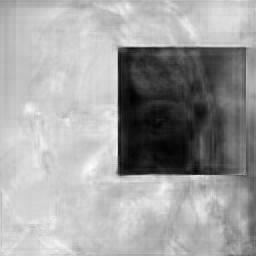} \end{minipage}\\

   \begin{minipage}{0.19\columnwidth} 	\hfill{} \begin{tabular}{c}	\footnotesize{} $D^U_{enc}$ class. \\
	\footnotesize{} score 	\end{tabular} 	\hfill{} \end{minipage} & 0.55 & 0.35 &  0.42 &  0.65 & & 0.35 & 0.60 & 0.85 & 0.76 \\

		 & & & & & \\
		 & & & & & \\
		 & & & & & \\

			 & \multicolumn{2}{c}{\footnotesize{} Real}  & \multicolumn{2}{c}{\footnotesize{} Fake} &  & \multicolumn{2}{c}{\footnotesize{} Real}  & \multicolumn{2}{c}{\footnotesize{} Fake}   \\
			  \vspace{0.5em}
		  \begin{minipage}{0.19\columnwidth} 	\hfill{} \begin{tabular}{c} {	\footnotesize{} Original} \\
		{\footnotesize{} images}\end{tabular} 	\hfill{} \end{minipage}  &
		 \begin{minipage}{0.2\columnwidth} 	\includegraphics[width=1\columnwidth, height=1\columnwidth]{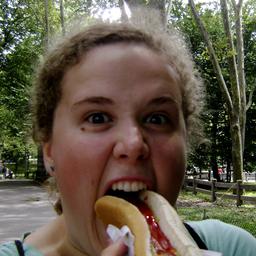}  \end{minipage} &
		  \begin{minipage}{0.2\columnwidth} 	\includegraphics[width=1\columnwidth, height=1\columnwidth]{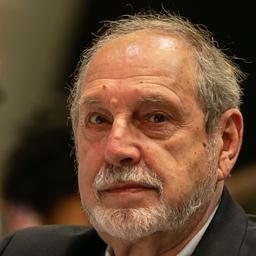}  \end{minipage} &
		 \begin{minipage}{0.2\columnwidth} 	\includegraphics[width=1\columnwidth, height=1\columnwidth]{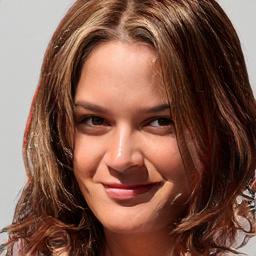}  \end{minipage} &
		 \begin{minipage}{0.2\columnwidth} 	\includegraphics[width=1\columnwidth, height=1\columnwidth]{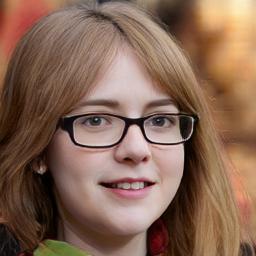}  \end{minipage} &
		 \begin{minipage}{0.1\columnwidth} $\ \ \ \ \ \ \ $ \end{minipage} &
		 \begin{minipage}{0.2\columnwidth} 	\includegraphics[width=1\columnwidth, height=1\columnwidth]{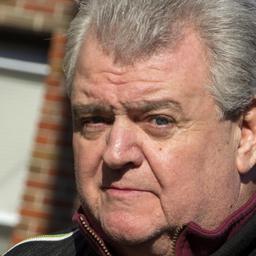}  \end{minipage} &
		 \begin{minipage}{0.2\columnwidth} 	\includegraphics[width=1\columnwidth, height=1\columnwidth]{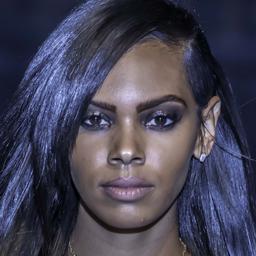}  \end{minipage} &
		 \begin{minipage}{0.2\columnwidth} 	\includegraphics[width=1\columnwidth, height=1\columnwidth]{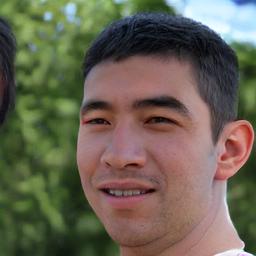}  \end{minipage} &
		 \begin{minipage}{0.2\columnwidth} 	\includegraphics[width=1\columnwidth, height=1\columnwidth]{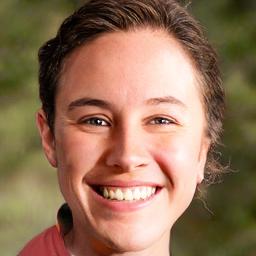}  \end{minipage}\\

\vspace{0.0em}
\begin{minipage}{0.19\columnwidth} 	\hfill{} \begin{tabular}{c}	\footnotesize{}Real/fake \\
		\footnotesize{} ratio $r$ 	\end{tabular} 	\hfill{} \end{minipage} & 0.36 & 0.56 &  0.93 &  0.79 & & 0.64 & 0.76 &  0.69 &  0.56 \\

\vspace{0.5em}

		\begin{minipage}{0.19\columnwidth}	\hfill{}	\begin{tabular}{c}	\footnotesize{} Mask \\
				\footnotesize{} $\mathrm{M}$ \end{tabular} 	\hfill{} \end{minipage} &
		\begin{minipage}{0.2\columnwidth}	\includegraphics[width=1\columnwidth, height=1\columnwidth, cfbox=verylightgray 0.2pt 0.2pt]{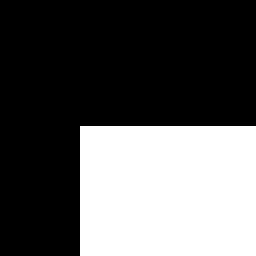} \end{minipage}&
		\begin{minipage}{0.2\columnwidth}	\includegraphics[width=1\columnwidth, height=1\columnwidth, cfbox=verylightgray 0.2pt 0.2pt]{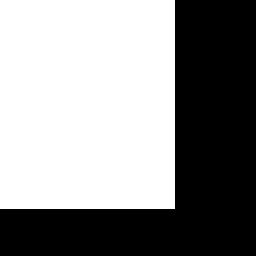} \end{minipage} &
		\begin{minipage}{0.2\columnwidth}	\includegraphics[width=1\columnwidth, height=1\columnwidth, cfbox=verylightgray 0.2pt 0.2pt]{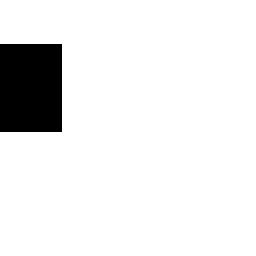} \end{minipage} &
		\begin{minipage}{0.2\columnwidth}	\includegraphics[width=1\columnwidth, height=1\columnwidth, cfbox=verylightgray 0.2pt 0.2pt]{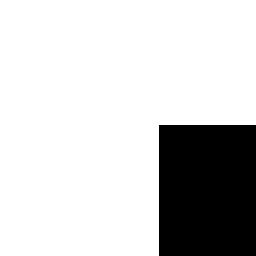} \end{minipage} &
		 \begin{minipage}{0.1\columnwidth} $\ \ \ \ \ \ \ $ \end{minipage} &
		\begin{minipage}{0.2\columnwidth}	\includegraphics[width=1\columnwidth, height=1\columnwidth, cfbox=verylightgray 0.2pt 0.2pt]{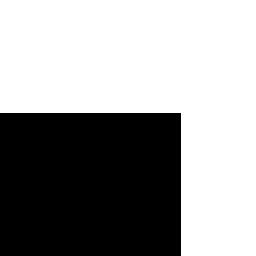} \end{minipage}&
		\begin{minipage}{0.2\columnwidth}	\includegraphics[width=1\columnwidth, height=1\columnwidth, cfbox=verylightgray 0.2pt 0.2pt]{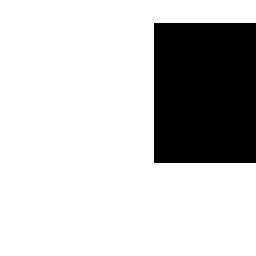} \end{minipage} &
		\begin{minipage}{0.2\columnwidth}	\includegraphics[width=1\columnwidth, height=1\columnwidth, cfbox=verylightgray 0.2pt 0.2pt]{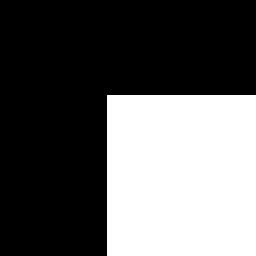} \end{minipage} &
		\begin{minipage}{0.2\columnwidth}	\includegraphics[width=1\columnwidth, height=1\columnwidth, cfbox=verylightgray 0.2pt 0.2pt]{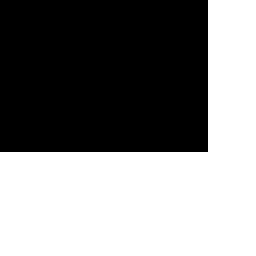} \end{minipage}\\
		\vspace{0.5em}

		  \begin{minipage}{0.19\columnwidth}	\hfill{}	\begin{tabular}{c}	\footnotesize{} CutMix \\
			\footnotesize{} images \end{tabular} 	\hfill{} \end{minipage} &
		  \begin{minipage}{0.2\columnwidth}	\includegraphics[width=1\columnwidth, height=1\columnwidth]{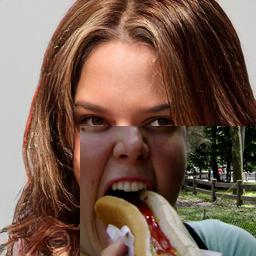} \end{minipage}&
		  \begin{minipage}{0.2\columnwidth}	\includegraphics[width=1\columnwidth, height=1\columnwidth]{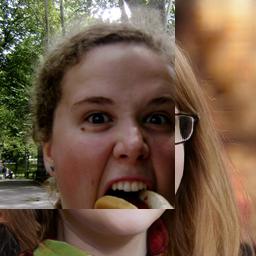} \end{minipage} &
		  \begin{minipage}{0.2\columnwidth}	\includegraphics[width=1\columnwidth, height=1\columnwidth]{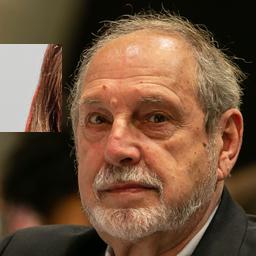} \end{minipage} &
		  \begin{minipage}{0.2\columnwidth}	\includegraphics[width=1\columnwidth, height=1\columnwidth]{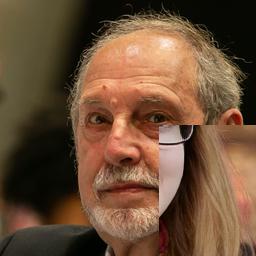} \end{minipage} &
		 \begin{minipage}{0.1\columnwidth} $\ \ \ \ \ \ \ $ \end{minipage} &
		 \begin{minipage}{0.2\columnwidth}	\includegraphics[width=1\columnwidth, height=1\columnwidth]{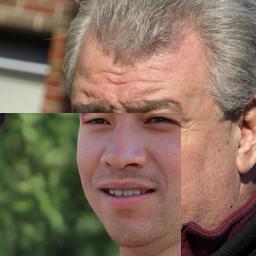} \end{minipage}&
		 \begin{minipage}{0.2\columnwidth}	\includegraphics[width=1\columnwidth, height=1\columnwidth]{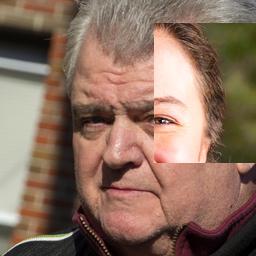} \end{minipage} &
		 \begin{minipage}{0.2\columnwidth}	\includegraphics[width=1\columnwidth, height=1\columnwidth]{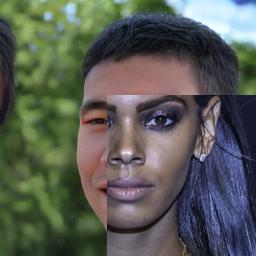} \end{minipage} &
		 \begin{minipage}{0.2\columnwidth}	\includegraphics[width=1\columnwidth, height=1\columnwidth]{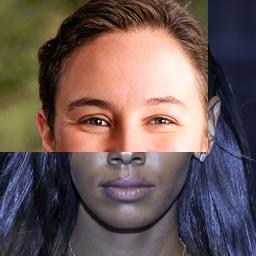} \end{minipage}\\
		 \vspace{0.5em}

		   \begin{minipage}{0.19\columnwidth} 	\hfill{} \begin{tabular}{c}	\footnotesize{} $D^U_{dec}$ segm. \\
			\footnotesize{} map 	\end{tabular} 	\hfill{} \end{minipage} &
		\begin{minipage}{0.2\columnwidth} \includegraphics[width=1\columnwidth, height=1\columnwidth]{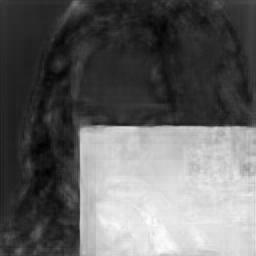} \end{minipage} &
		\begin{minipage}{0.2\columnwidth} \includegraphics[width=1\columnwidth, height=1\columnwidth]{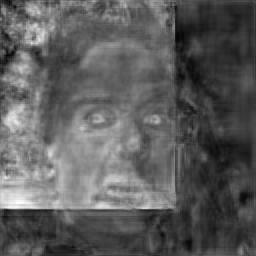} \end{minipage} &
		\begin{minipage}{0.2\columnwidth} \includegraphics[width=1\columnwidth, height=1\columnwidth]{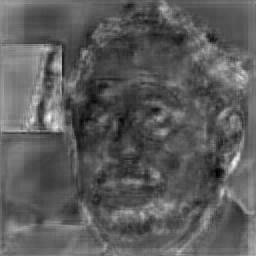}  \end{minipage} &
		\begin{minipage}{0.2\columnwidth} \includegraphics[width=1\columnwidth, height=1\columnwidth]{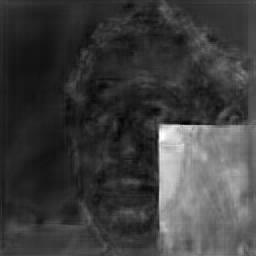} \end{minipage} &
		 \begin{minipage}{0.1\columnwidth} $\ \ \ \ \ \ \ $ \end{minipage} &
		\begin{minipage}{0.2\columnwidth} \includegraphics[width=1\columnwidth, height=1\columnwidth]{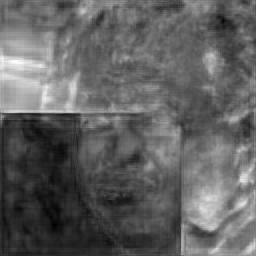} \end{minipage} &
		\begin{minipage}{0.2\columnwidth} \includegraphics[width=1\columnwidth, height=1\columnwidth]{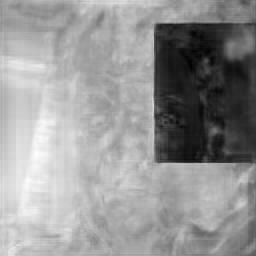} \end{minipage} &
		\begin{minipage}{0.2\columnwidth} \includegraphics[width=1\columnwidth, height=1\columnwidth]{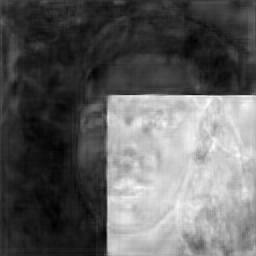}  \end{minipage} &
		\begin{minipage}{0.2\columnwidth} \includegraphics[width=1\columnwidth, height=1\columnwidth]{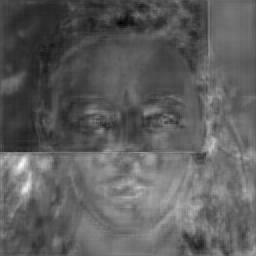} \end{minipage}   \\

		   \begin{minipage}{0.19\columnwidth} 	\hfill{} \begin{tabular}{c}	\footnotesize{} $D^U_{enc}$ class. \\
			\footnotesize{} score 	\end{tabular} 	\hfill{} \end{minipage} & 0.51 & 0.47 &  0.26 &  0.21 & & 0.30 & 0.82 & 0.33 & 0.41\\

				\end{tabular}

	\par\end{centering}

	\vspace{0.5em}

	\caption{\label{fig:supp_cutmix}  Visualization of the CutMix augmentation and the predictions of the U-Net discriminator on CutMix images. 1st row: real and fake samples. 2nd\&3rd rows: sampled real/fake CutMix ratio $r$ and corresponding binary masks $\mathrm{M}$ (color code: white for real, black for fake). 4th row: generated CutMix images from real and fake samples. 5th\&6th row: the corresponding real/fake segmentation maps of the U-Net GAN decoder $D^U_{dec}$ with the corresponding predicted classification scores by the encoder $D^U_{enc}$ below. 
	 }
\end{figure*}

In Figure \ref{fig:supp_cutmix} we show more examples of the CutMix images and the corresponding U-Net based discriminator $D^U$ predictions. Note that in many cases, the decoder output for fake image patches is darker than for real image ones. However, the predicted intensity for an identical local patch can change for different mixing scenarios. This indicates that the U-Net discriminator takes contextual information into account for local decisions.

\clearpage
\subsection{Qualitative Results on COCO-Animals}\label{sec:syn samples_coco}

Here we present more qualitative results of U-Net GAN on COCO-Animals~\cite{Lin2014MicrosoftCC,OpenImages}.
We use COCO-Animals for class conditional image synthesis and generate images with the resolution of $128\times 128$.

\subsubsection*{Generated COCO-Animals samples}
\begin{figure*}
\begin{centering}
\setlength{\tabcolsep}{0.0em}
\renewcommand{\arraystretch}{1.0}
\par\end{centering}
\begin{centering}
\hfill{}%
\begin{tabular}{@{\hskip -0.1in}c@{\hskip 0.03in}c@{\hskip 0.03in}c@{\hskip 0.03in}c@{\hskip 0.03in}c@{\hskip 0.03in}c@{\hskip 0.03in}c@{\hskip 0.03in}c@{\hskip 0.03in}c@{\hskip 0.03in}c@{}}
\includegraphics[width=0.1\textwidth]{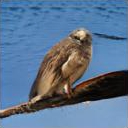} &
\includegraphics[width=0.1\textwidth]{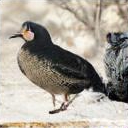} &
\includegraphics[width=0.1\textwidth]{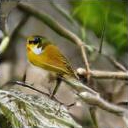} &
\includegraphics[width=0.1\textwidth]{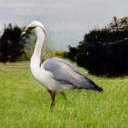} &
\includegraphics[width=0.1\textwidth]{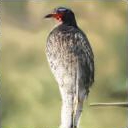}  & %
\includegraphics[width=0.1\textwidth]{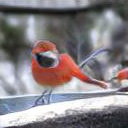}  &
\includegraphics[width=0.1\textwidth]{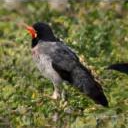} &
\includegraphics[width=0.1\textwidth]{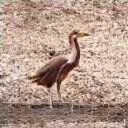}  & %
\includegraphics[width=0.1\textwidth]{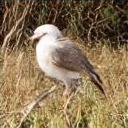}  &
\includegraphics[width=0.1\textwidth]{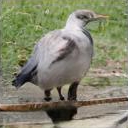}
\tabularnewline

\includegraphics[width=0.1\textwidth]{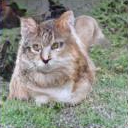} &
\includegraphics[width=0.1\textwidth]{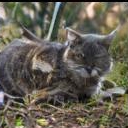} &
\includegraphics[width=0.1\textwidth]{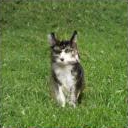} &
\includegraphics[width=0.1\textwidth]{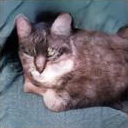}  &
\includegraphics[width=0.1\textwidth]{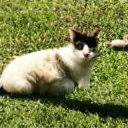}  & %
\includegraphics[width=0.1\textwidth]{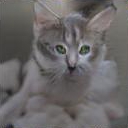}  &
\includegraphics[width=0.1\textwidth]{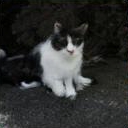} &
\includegraphics[width=0.1\textwidth]{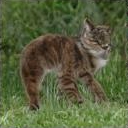}  & %
\includegraphics[width=0.1\textwidth]{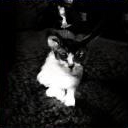}  &
\includegraphics[width=0.1\textwidth]{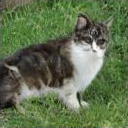}
\tabularnewline

\includegraphics[width=0.1\textwidth]{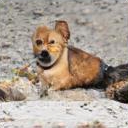} &
\includegraphics[width=0.1\textwidth]{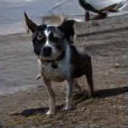} &
\includegraphics[width=0.1\textwidth]{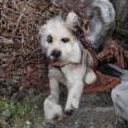} &
\includegraphics[width=0.1\textwidth]{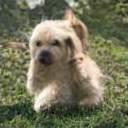} &
\includegraphics[width=0.1\textwidth]{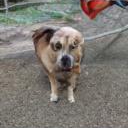}  & %
\includegraphics[width=0.1\textwidth]{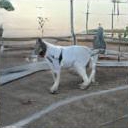}  &
\includegraphics[width=0.1\textwidth]{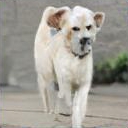} &
\includegraphics[width=0.1\textwidth]{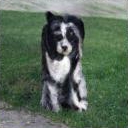}  & %
\includegraphics[width=0.1\textwidth]{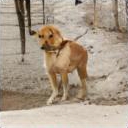}  &
\includegraphics[width=0.1\textwidth]{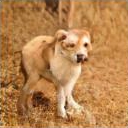}
\tabularnewline

\includegraphics[width=0.1\textwidth]{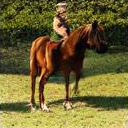} &
\includegraphics[width=0.1\textwidth]{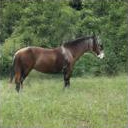} &
\includegraphics[width=0.1\textwidth]{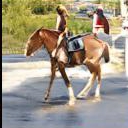} &
\includegraphics[width=0.1\textwidth]{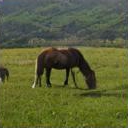}  &
\includegraphics[width=0.1\textwidth]{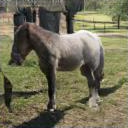}  & %
\includegraphics[width=0.1\textwidth]{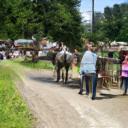}  &
\includegraphics[width=0.1\textwidth]{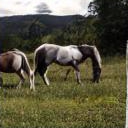} &
\includegraphics[width=0.1\textwidth]{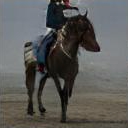}  & %
\includegraphics[width=0.1\textwidth]{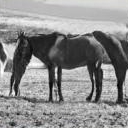}  &
\includegraphics[width=0.1\textwidth]{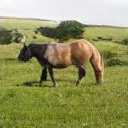}
\tabularnewline

\includegraphics[width=0.1\textwidth]{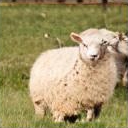} &
\includegraphics[width=0.1\textwidth]{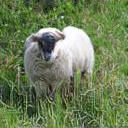} &
\includegraphics[width=0.1\textwidth]{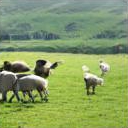} &
\includegraphics[width=0.1\textwidth]{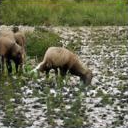} &
\includegraphics[width=0.1\textwidth]{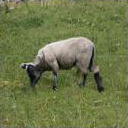}  & %
\includegraphics[width=0.1\textwidth]{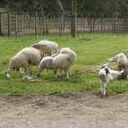}  &
\includegraphics[width=0.1\textwidth]{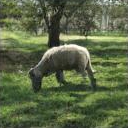} &
\includegraphics[width=0.1\textwidth]{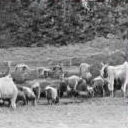}  & %
\includegraphics[width=0.1\textwidth]{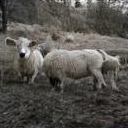}  &
\includegraphics[width=0.1\textwidth]{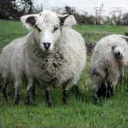}
\tabularnewline

\includegraphics[width=0.1\textwidth]{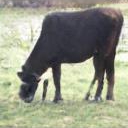} &
\includegraphics[width=0.1\textwidth]{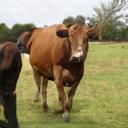} &
\includegraphics[width=0.1\textwidth]{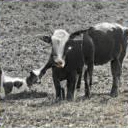} &
\includegraphics[width=0.1\textwidth]{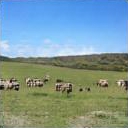}  &
\includegraphics[width=0.1\textwidth]{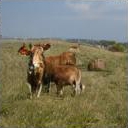}  & %
\includegraphics[width=0.1\textwidth]{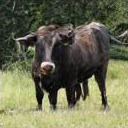}  &
\includegraphics[width=0.1\textwidth]{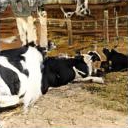} &
\includegraphics[width=0.1\textwidth]{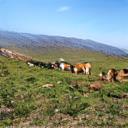}  & %
\includegraphics[width=0.1\textwidth]{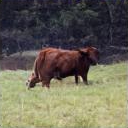}  &
\includegraphics[width=0.1\textwidth]{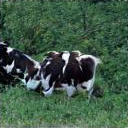}
\tabularnewline

\includegraphics[width=0.1\textwidth]{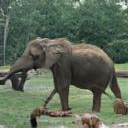} &
\includegraphics[width=0.1\textwidth]{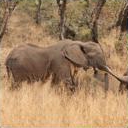} &
\includegraphics[width=0.1\textwidth]{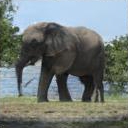} &
\includegraphics[width=0.1\textwidth]{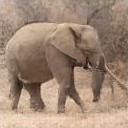} &
\includegraphics[width=0.1\textwidth]{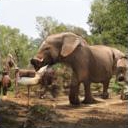}  & %
\includegraphics[width=0.1\textwidth]{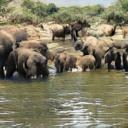}  &
\includegraphics[width=0.1\textwidth]{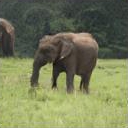} &
\includegraphics[width=0.1\textwidth]{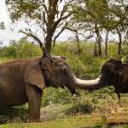}  & %
\includegraphics[width=0.1\textwidth]{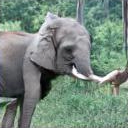}  &
\includegraphics[width=0.1\textwidth]{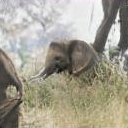}
\tabularnewline

\includegraphics[width=0.1\textwidth]{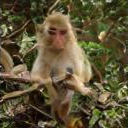} &
\includegraphics[width=0.1\textwidth]{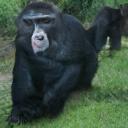} &
\includegraphics[width=0.1\textwidth]{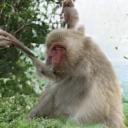} &
\includegraphics[width=0.1\textwidth]{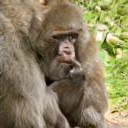}  &
\includegraphics[width=0.1\textwidth]{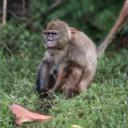}  & %
\includegraphics[width=0.1\textwidth]{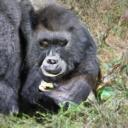}  &
\includegraphics[width=0.1\textwidth]{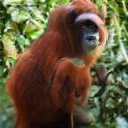} &
\includegraphics[width=0.1\textwidth]{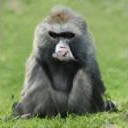}  & %
\includegraphics[width=0.1\textwidth]{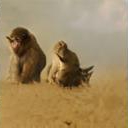}  &
\includegraphics[width=0.1\textwidth]{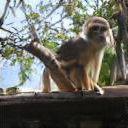}
\tabularnewline

\includegraphics[width=0.1\textwidth]{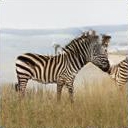} &
\includegraphics[width=0.1\textwidth]{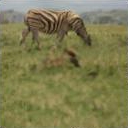} &
\includegraphics[width=0.1\textwidth]{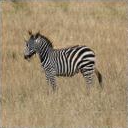} &
\includegraphics[width=0.1\textwidth]{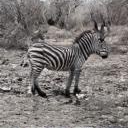} &
\includegraphics[width=0.1\textwidth]{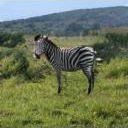}  & %
\includegraphics[width=0.1\textwidth]{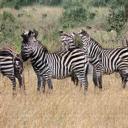}  &
\includegraphics[width=0.1\textwidth]{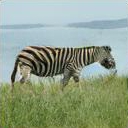} &
\includegraphics[width=0.1\textwidth]{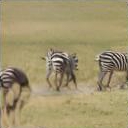}  & %
\includegraphics[width=0.1\textwidth]{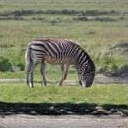}  &
\includegraphics[width=0.1\textwidth]{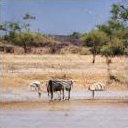}
\tabularnewline

\includegraphics[width=0.1\textwidth]{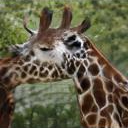} &
\includegraphics[width=0.1\textwidth]{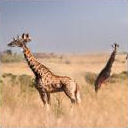} &
\includegraphics[width=0.1\textwidth]{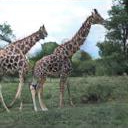} &
\includegraphics[width=0.1\textwidth]{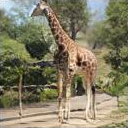}  &
\includegraphics[width=0.1\textwidth]{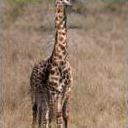}  & %
\includegraphics[width=0.1\textwidth]{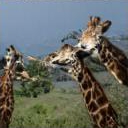}  &
\includegraphics[width=0.1\textwidth]{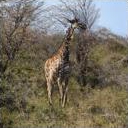} &
\includegraphics[width=0.1\textwidth]{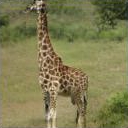}  & %
\includegraphics[width=0.1\textwidth]{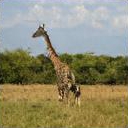}  &
\includegraphics[width=0.1\textwidth]{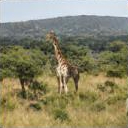}
\tabularnewline

\end{tabular}\hfill{}
\par\end{centering}
\caption{\label{fig:coco_pics2} Images generated with U-Net GAN trained on COCO-Animals with resolution $128 \times 128$.}
 
\end{figure*}

Figure \ref{fig:coco_pics2} shows generated samples of different classes on COCO-Animals. We observe images of good quality and high intra-class variation. We further notice that employing the class-conditional projection (as used in BigGAN) in the pixel output space of the decoder does not introduce class leakage or influence the class separation in any other way. 
These observations further confirm that our U-Net GAN is effective in class-conditional image generation as well.

\subsubsection*{Per-pixel U-Net discriminator feedback}

\begin{figure*}
	\begin{centering}
		\setlength{\tabcolsep}{0.2em}
		\renewcommand{\arraystretch}{1.0}
		\par\end{centering}
	\begin{centering}
		\begin{tabular}{@{}c@{\hskip 0.05in }c@{\hskip 0.05in }c@{\hskip 0.05in }c@{\hskip 0.05in }c@{\hskip 0.05in }c@{\hskip 0.05in }c@{\hskip 0.05in }c@{\hskip 0.05in }c@{}}

			{\footnotesize{}}\includegraphics[width=0.12\textwidth]{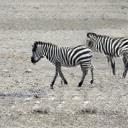} &
			{\footnotesize{}}\includegraphics[width=0.12\textwidth]{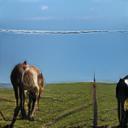} &
			{\footnotesize{}}\includegraphics[width=0.12\textwidth]{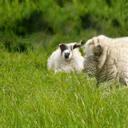} &
			{\footnotesize{}}\includegraphics[width=0.12\textwidth]{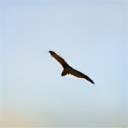} &
			{\footnotesize{}}\includegraphics[width=0.12\textwidth]{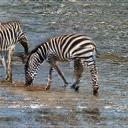} &
			{\footnotesize{}}\includegraphics[width=0.12\textwidth]{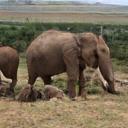} &
			{\footnotesize{}}\includegraphics[width=0.12\textwidth]{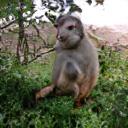} &
			{\footnotesize{}}\includegraphics[width=0.12\textwidth]{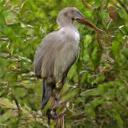} \\

			{\footnotesize{}}\includegraphics[width=0.12\textwidth]{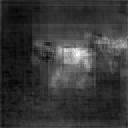} &
			{\footnotesize{}}\includegraphics[width=0.12\textwidth]{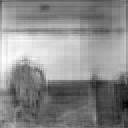} &
			{\footnotesize{}}\includegraphics[width=0.12\textwidth]{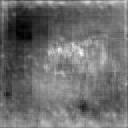} &
			{\footnotesize{}}\includegraphics[width=0.12\textwidth]{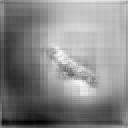} &
			{\footnotesize{}}\includegraphics[width=0.12\textwidth]{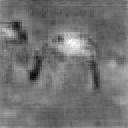} &
			{\footnotesize{}}\includegraphics[width=0.12\textwidth]{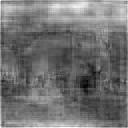} &
			{\footnotesize{}}\includegraphics[width=0.12\textwidth]{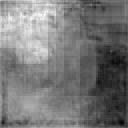} &
			{\footnotesize{}}\includegraphics[width=0.12\textwidth]{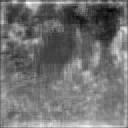} \\

		\end{tabular}

		\par\end{centering}
	\caption{\label{fig:coco_response}  Generated samples on COCO-Animals and the corresponding U-Net decoder predictions. Brighter colors correspond to the discriminator confidence of pixel being real (and darker of being fake).}
\end{figure*}

Figure \ref{fig:coco_response} shows generated examples and the corresponding per-pixel predictions of the U-Net discriminator. We observe that the resulting maps often tend to exhibit a bias towards objects.

\subsubsection*{Interpolations in the latent space}

\begin{figure*}
\begin{centering}
\setlength{\tabcolsep}{0.1em}
\renewcommand{\arraystretch}{1.0}
\par\end{centering}
\begin{centering}

\begin{tabular}{@{}c@{}}

\includegraphics[width=1\textwidth]{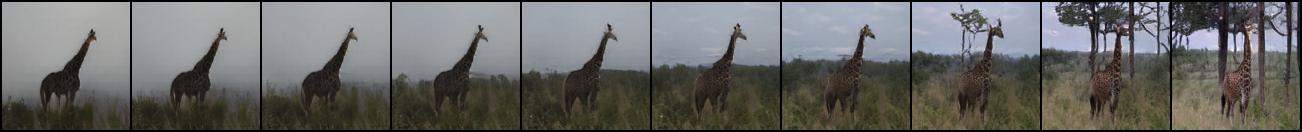}\tabularnewline

\includegraphics[width=1\textwidth]{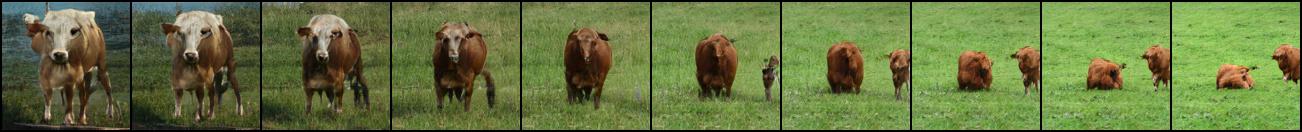}\tabularnewline

\includegraphics[width=1\textwidth]{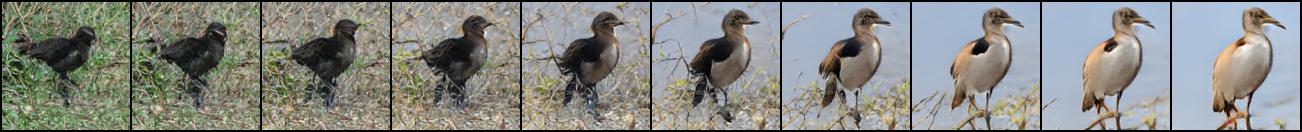}\tabularnewline

\includegraphics[width=1\textwidth]{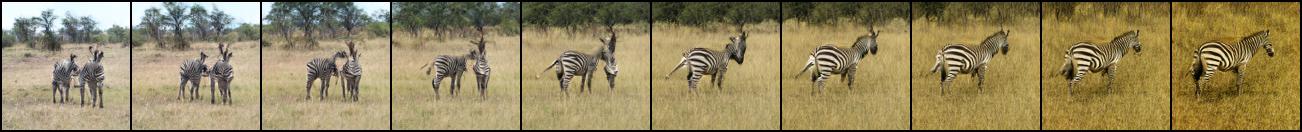}\tabularnewline

\includegraphics[width=1\textwidth]{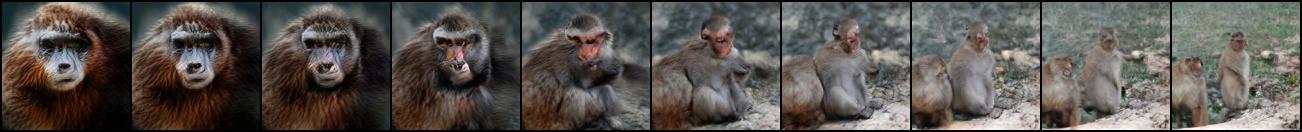}\tabularnewline

\includegraphics[width=1\textwidth]{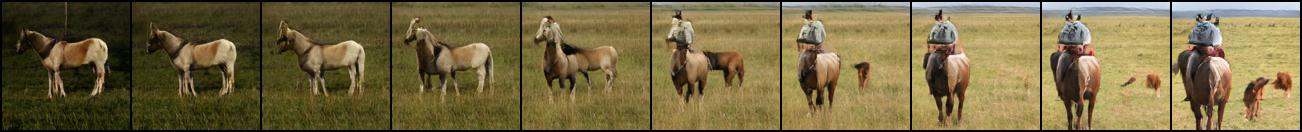}\tabularnewline

\includegraphics[width=1\textwidth]{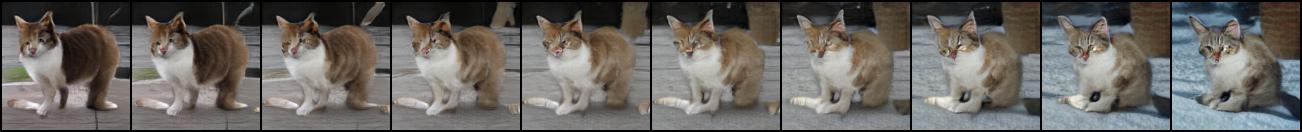}\tabularnewline

\includegraphics[width=1\textwidth]{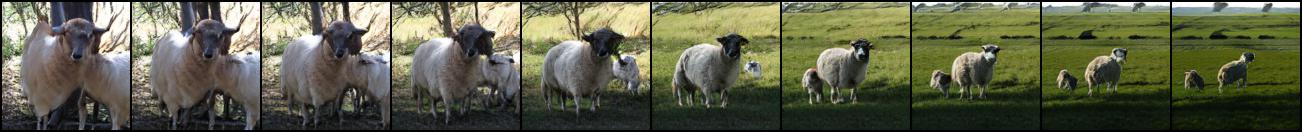}\tabularnewline

\includegraphics[width=1\textwidth]{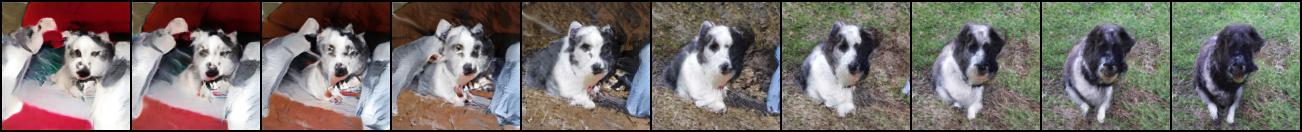}\tabularnewline

\includegraphics[width=1\textwidth]{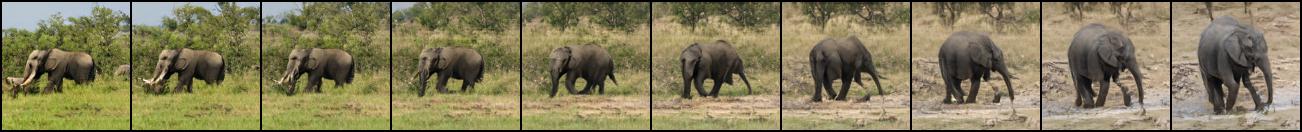}\tabularnewline

\end{tabular}
\par\end{centering}
\caption{\label{fig:more_coco_interpolations} Images generated with U-Net GAN on COCO-Animals with resolution $128 \times 128$ when interpolating in the latent space between two synthetic samples (left to right).} 
\vspace{0em}
\end{figure*}

Figure~\ref{fig:more_coco_interpolations} displays images generated on COCO-Animals by U-Net GAN through linear interpolation in the latent space between two synthetic samples. 
We observe that the interpolations are semantically smooth between different classes of animals, e.g. background seamlessly changes between two scenes, number of instances gradually increases or decreases, shape and color of objects smoothly changes from left to right. 

\subsection{Details on the COCO-Animals Dataset}

COCO-Animals is a medium-sized ($\sim 38k$) dataset composed of 10 animal classes, and is intended for experiments that demand a high-resolution equivalent for \mbox{CIFAR10}.
The categories are \textit{bird, cat, dog, horse, cow, sheep, giraffe, zebra, elephant, and monkey}. The images are taken from COCO~\cite{Lin2014MicrosoftCC} and the OpenImages~\cite{OpenImages} subset that provides semantic label maps and binary mask and is also human-verified. The two datasets have a great overlap in animal classes. We take \textit{all} images from COCO and the aforementioned OpenImages split in the categories \textit{horse, cow, sheep, giraffe, zebra} and \textit{elephant}. The \textit{monkey} images are taken over directly from OpenImages, since this category contained more training samples than the next biggest COCO animal class \textit{bear}. The class \textit{bear} and \textit{monkey} are not shared between COCO and OpenImages. Lastly, the categories \textit{bird, cat} and \textit{dog} contained vastly more samples than all other categories. For this reason, we took over only a subset of the total of all images in these categories. These samples were picked from OpenImages only, for their better visual quality. To ensure good quality of the picked examples, we used the provided bounding boxes to filter out images in which the animal of interest is either too small or too big ($>80\%,<30\%$ of the image area for cats, $>70\%,<50\%$ for birds and dogs). The thresholds were chose such that the number of appropriate images is approximately equal. %

\subsection{Architectures and Training Details}\label{sec:networks}

\begin{table*}[t]
	\vspace{-1em}
	\setlength{\tabcolsep}{0.25em}
	\renewcommand{\arraystretch}{1.5}
	\centering

	\begin{minipage}[t]{.45\textwidth}
	\centering
	\subcaption{BigGAN Generator ($128\times 128$, class-conditional)}
	\begin{tabular}{c}
	\hline 	\hline
    $z \in \mathbb{R}^{120}  \sim \mathcal{N}(0,I)$ \\
	Embed(y) $\in \mathbb{R}^{128}$\\
	\hline
	Linear (20 + 128) $\rightarrow 4 \times 4 \times 16ch$ \\
	\hline
	ResBlock up $16ch \rightarrow  16ch$ \\
	\hline
	ResBlock up $16ch \rightarrow  8ch$ \\
	\hline
	ResBlock up $8ch \rightarrow  4ch$ \\
	\hline
	ResBlock up $4ch \rightarrow  2ch$ \\
	\hline
	Non-Local Block $(64 \times 64)$ \\
	\hline
	ResBlock up $2ch \rightarrow  ch$ \\
	\hline
	BN, ReLU, $3 \times 3$ Conv $ch \rightarrow 3 $ \\
	\hline
	Tanh \\
	\hline 	\hline \\ \\
     \end{tabular}
\end{minipage} \hspace{0.5cm}	\begin{minipage}[t]{.45\textwidth}
\centering
\subcaption{BigGAN Discriminator ($128\times 128$, class-conditional)}
\begin{tabular}{c}
	\hline 	\hline
	RGB image $x \in \mathbb{R}^{128 \times 128 \times 3}$ \\
	\hline
	ResBlock down $ch \rightarrow  2ch$ \\
	\hline
	Non-Local Block $(64 \times 64)$ \\
	\hline
	ResBlock down $2ch \rightarrow  4ch$ \\
	\hline
	ResBlock down $4ch \rightarrow  8ch$ \\
	\hline
	ResBlock down $8ch \rightarrow  16ch$ \\
	\hline
	ResBlock down $16ch \rightarrow 16ch$ \\
	\hline
	ReLU, Global sum pooling \\
	\hline
	Embed(y)$\cdot h$ + (linear$\rightarrow 1$)	 \\
	\hline 	\hline \\ \\
\end{tabular}%
\end{minipage}

	\begin{minipage}[t]{.45\textwidth}
	\centering
	\subcaption{BigGAN Generator ($256\times 256$, unconditional)}
	\begin{tabular}{c}
		\hline 	\hline
		$z \in \mathbb{R}^{140} \sim \mathcal{N}(0,I)$ \\
		\hline
		Linear (20 + 128) $\rightarrow 4 \times 4 \times 16ch$ \\
		\hline
		ResBlock up $16ch \rightarrow  16ch$ \\
		\hline
		ResBlock up $16ch \rightarrow  8ch$ \\
		\hline
		ResBlock up $8ch \rightarrow  8ch$ \\
		\hline
		ResBlock up $8ch \rightarrow  4ch$ \\
		\hline
		ResBlock up $4ch \rightarrow  2ch$ \\
		\hline
		Non-Local Block $(128 \times 128)$ \\
		\hline
		ResBlock up $2ch \rightarrow  ch$ \\
		\hline
		BN, ReLU, $3 \times 3$ Conv $ch \rightarrow 3 $ \\
		\hline
		Tanh \\
		\hline 	\hline
	\end{tabular}
\end{minipage}  \hspace{0.5cm}
	\begin{minipage}[t]{.45\textwidth}
	\centering
	\subcaption{BigGAN Discriminator ($256\times 256$, unconditional)}
	\begin{tabular}{c}
	\hline 	\hline
    RGB image $x \in \mathbb{R}^{256 \times 256 \times 3}$ \\
	\hline
	ResBlock down $ch \rightarrow  2ch$ \\
	\hline
	ResBlock down $2ch \rightarrow  4ch$ \\
	\hline
	Non-Local Block $(64 \times 64)$ \\
	\hline
	ResBlock down $4ch \rightarrow  8ch$ \\
	\hline
	ResBlock down $8ch \rightarrow  8ch$ \\
	\hline
	ResBlock down $8ch \rightarrow  16ch$ \\
	\hline
	ResBlock down $16ch \rightarrow 16ch$ \\
	\hline
	ReLU, Global sum pooling \\
	\hline
	linear$\rightarrow 1$\\
	\hline 	\hline
     \end{tabular}
\end{minipage} \hspace{0.5cm}

    \caption{The BigGAN~\cite{Brock2019} generator and discriminator architectures for class-conditional and unconditional tasks of generating images at different resolutions. Top (a and b): The class-conditional BigGAN model for resolution $128\times128$. Bottom (c and d): The BigGAN model for resolution $256\times 256$, modified to be \textit{un}conditional.} \label{table:supp_biggan_arch} %
\end{table*}

\begin{table*}%
	\vspace{0em}
	\setlength{\tabcolsep}{0.25em}
	\renewcommand{\arraystretch}{1.5}
	\centering

	\begin{minipage}[t]{.45\textwidth}
			\centering
			\subcaption{U-Net GAN Discriminator  ($256\times 256$, unconditional)}
	\begin{tabular}{c}
	
	\hline 	\hline
    RGB image $x \in \mathbb{R}^{256 \times 256 \times 3}$ \\
	\hline
	ResBlock down $ch \rightarrow  2ch$ \\
	\hline
	ResBlock down $2ch \rightarrow  4ch$ \\
	\hline
	Optional Non-Local Block $(64 \times 64)$ \\
	\hline
	ResBlock down $4ch \rightarrow  8ch$ \\
	\hline
	ResBlock down $8ch \rightarrow  8ch$ \\
	\hline
	ResBlock down $8ch \rightarrow  16ch$ *(see below) \\
	\hline
	\hline
	ResBlock up $16ch \rightarrow  8ch$ \\
	\hline
	ResBlock up $(8+8)ch \rightarrow  8ch$ \\
	\hline
	ResBlock up $(8+8)ch \rightarrow  4ch$ \\
	\hline
	ResBlock up $(4+4)ch \rightarrow  2ch$ \\
	\hline
	ResBlock up $(2+2)ch \rightarrow  ch$ \\
	\hline
	ResBlock up $(ch + ch) \rightarrow  ch$ \\
	\hline
	ResBlock $ch \rightarrow 1 $ \\
	\hline
	Sigmoid \\
	\hline 	\hline 
	* ReLU, Global sum pooling, linear$\rightarrow 1$	 \\
 \hline 	\hline
   \end{tabular}%
 
\end{minipage} \hspace{0.5cm}
	\begin{minipage}[t]{.45\textwidth}	
			\centering
			  			\subcaption{U-Net GAN Discriminator($128\times 128$, class-conditional)}
\begin{tabular}{c}

	\hline 	\hline
    RGB image $x \in \mathbb{R}^{128 \times 128 \times 3}$ \\
	\hline
	ResBlock down $ch \rightarrow  2ch$ \\
	\hline
	Optional Non-Local Block $(64 \times 64)$ \\
	\hline
	ResBlock down $2ch \rightarrow  4ch$ \\
	\hline
	ResBlock down $8ch \rightarrow  8ch$ \\
	\hline
	ResBlock down $8ch \rightarrow  16ch$ *(see below) \\
	\hline
	\hline
	ResBlock up $16ch \rightarrow  8ch$ \\
	\hline
	ResBlock up $(8+8)ch \rightarrow  4ch$ \\
	\hline
	ResBlock up $(4+4)ch \rightarrow  2ch$ \\
	\hline
	ResBlock up $(2+2)ch \rightarrow  ch$ \\
	\hline
	ResBlock up $(ch + ch) \rightarrow  ch$ \\
	\hline
	Embed(y)$\cdot h$ + (Conv $ch\rightarrow 1$)	 \\
	\hline
	Sigmoid \\
	\hline 	\hline
	* ReLU, Global sum pooling \\
	\hline
	Embed(y)$\cdot h$ + (linear$\rightarrow 1$)	 	 \\
 \hline 	\hline 
     \end{tabular}%

	\end{minipage}	
\caption{The U-Net GAN discriminator architectures for class-conditional (a) and unconditional (b) tasks of generating images at resolution $128\times128$ and $256\times 256$, respectively.} \label{table:supp_unet_arch} %
\end{table*}

Architecture details of the BigGAN model~\cite{Brock2019} and our U-Net discriminator are summarized in Table \ref{table:supp_biggan_arch} and Table \ref{table:supp_unet_arch}. From these tables it is easy to see that the encoder and decoder of the U-Net discriminator follow the original BigGAN discriminator and generator setups, respectively. One difference is that the number of input channels in the U-Net decoder is doubled, since encoder features are concatenated to the input features. 

Table \ref{table:supp_unet_arch} presents two U-Net discriminator networks: a class-conditional discriminator for image resolution $128\times 128$, and an unconditional discriminator for resolution $256\times 256$. The decoder does not have $3$ output channels (like the BigGAN generator that it is copied from), but $ch=64$ channels, resulting in a feature map $h$ of size $64\times 128 \times 128$, to which a $1\times 1$ convolution is applied to reduce the number of channels to $1$. In the class-conditional architecture, a learned class-embedding is multiplied with the aforementioned $64$-dimensional output $h$ at every spatial position, and summed along the channel dimension (corresponding to the inner product). The resulting map of size $1\times 128 \times 128$ is added to the output, leaving us with $128 \times 128$ logits.

We follow~\cite{Brock2019} for setting up the hyperparameters for training U-Net GAN, which are summarized in Table \ref{table:hp}.
\begin{table}[H]
\begin{center}
\begin{tabular}{ |l|c| } 
 \hline
 Hyperparameter & Value \\ 
 \hline
 Optimizer & Adam ($\beta_1=0, \beta_2=0.999$)  \\ 
 G's learning rate  & 1e-4 ($256\time256$), 5e-5 ($128\time128$)  \\ 
 D's learning rate  & 5e-4 ($256\time256$), 2e-4 ($128\time128$) \\ 
 Batch size & 20 ($256\time256$), 80 ($128\time128$) \\ 
 Weight Initialization & Orthogonal \\
\hline
\end{tabular}
\end{center}
\caption{Hyperparameters of U-Net GAN}\label{table:hp}
\end{table}

Regarding the difference between class-conditional and unconditional image generation, it is worth noting that the CutMix regularization is applied only to samples within the same class. In other words, real and generated samples are mixed only within the class (e.g. real and fake zebras, but not real zebras with fake elephants).

\end{document}